\RequirePackage{amsthm}
\documentclass[sn-mathphys-num, iicol]{sn-jnl}

\usepackage{graphicx}%
\usepackage{multirow}%
\usepackage{amsmath,amssymb,amsfonts}%
\usepackage{amsthm}%
\usepackage{mathrsfs}%
\usepackage[title]{appendix}%
\usepackage{xcolor}%
\usepackage{textcomp}%
\usepackage{manyfoot}%
\usepackage{booktabs}%
\usepackage{algorithm}%
\usepackage{algorithmicx}%
\usepackage{algpseudocode}%
\usepackage{listings}%
\usepackage{subfigure}
\usepackage{float}
\usepackage{pifont}
\newcommand{\cmark}{\ding{51}}%
\newcommand{\xmark}{\ding{55}}%
\usepackage{lmodern}
\usepackage{anyfontsize}
\usepackage{caption}
\usepackage{geometry} 
\theoremstyle{thmstyleone}%
\theoremstyle{thmstyletwo}%

\theoremstyle{thmstylethree}%

\begin{document}

\title[Article Title]{Graph in Graph Neural Network}


\author[1]{\fnm{Jiongshu} \sur{Wang}}\email{jiongshuwang@gmail.com}
\author[3]{\fnm{Jing} \sur{Yang}}\email{y.jing2016@gmail.com}
\author[2]{\fnm{Jiankang} \sur{Deng}}\email{jiankangdeng@gmail.com}
\author[3]{\fnm{Hatice} \sur{Gunes}}\email{Hatice.Gunes@cl.cam.ac.uk}
\author*[1,3]{\fnm{Siyang} \sur{Song}}\email{ss2796@cam.ac.uk}

\affil*[1]{\orgdiv{School of Computing and Mathematical Sciences}, \orgname{University of Leicester}, \orgaddress{\city{Leicester}, \postcode{LE1 7RH}, \state{East Midlands}, \country{United Kingdom}}}

\affil[2]{\orgdiv{Department of Computing}, \orgname{Imperial College London}, \orgaddress{\city{London}, \postcode{SW7 2AZ}, \state{London}, \country{United Kingdom}}}

\affil*[3]{\orgdiv{Department of Computer Science and Technology}, \orgname{University of Cambridge}, \orgaddress{\city{Cambridge}, \postcode{CB3 0FD}, \state{Cambridgeshire}, \country{United Kingdom}}}

\abstract{Existing Graph Neural Networks (GNNs) are limited to process graphs each of whose vertices is represented by a vector or a single value, limited their representing capability to describe complex objects. In this paper, we propose \textbf{the first GNN (called Graph in Graph Neural (GIG) Network)} which can process graph-style data (called GIG sample) whose vertices are further represented by graphs. Given a set of graphs or a data sample whose components can be represented by a set of graphs (called multi-graph data sample), our GIG network starts with a GIG sample generation (GSG) module which encodes the input as a \textbf{GIG sample}, where each GIG vertex includes a graph. Then, a set of GIG hidden layers are stacked, with each consisting of: (1) a \textbf{GIG vertex-level updating (GVU)} module that individually updates the graph in every GIG vertex based on its internal information; and (2) a \textbf{global-level GIG sample updating (GGU)} module that updates graphs in all GIG vertices based on their relationships, making the updated GIG vertices become global context-aware. This way, both internal cues within the graph contained in each GIG vertex and the relationships among GIG vertices could be utilized for down-stream tasks. Experimental results demonstrate that our GIG network generalizes well for not only various generic graph analysis tasks but also real-world multi-graph data analysis (e.g., human skeleton video-based action recognition), which achieved the new state-of-the-art results on 13 out of 14 evaluated datasets. Our code is publicly available at \url{https://github.com/wangjs96/Graph-in-Graph-Neural-Network}.}




\maketitle

\section{Introduction}
\label{sec1}

Graph Neural Networks (GNNs)~\cite{west2001introduction} are deep neural networks specifically designed to process graph-structured data. Recent advances in GNNs (e.g.,  Graphormer \cite{ying2106transformers}, EGNN \cite{satorras2021n}, SIGN \cite{frasca2020sign} and GatedGCN \cite{bresson2017residual}) can effectively extract task-specific features from non-Euclidean graphs (e.g., Graph data, Tree structures, Manifolds and Mesh networks), where each graph vertex describes the mathematical abstraction of an object \cite{hamilton2020graph}. Consequently, GNNs have been successfully developed for a diverse range of real-world applications such as social network analysis \cite{pareja2020evolvegcn,song2024merg}, drug discovery \cite{yang2019analyzing}, recommendation systems \cite{he2020lightgcn,liu2023smef}, object recognition and segmentation \cite{brissman2023recurrent,wang2024accurate,wu2024domain,deng2024advancing}, and human behaviour understanding \cite{yan2018spatial,song2022gratis,ma2021deep,tu2024dual,qiao2023joint,song2022learning,xu2024two}.

\begin{figure}[ht]
\begin{center}
\includegraphics[width=.45\textwidth]{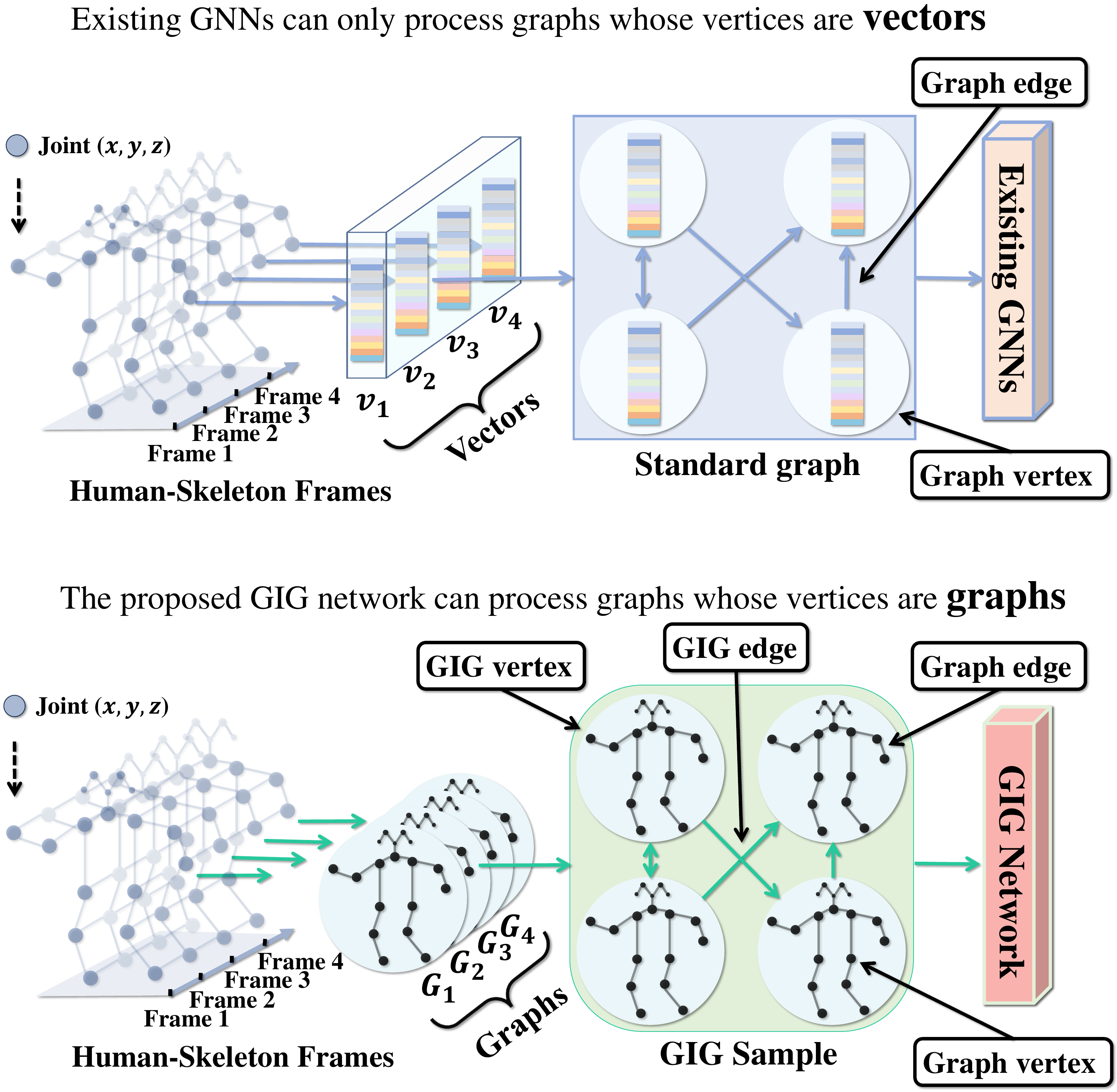}
\end{center}
   \caption{Comparison between existing GNNs and our GIG network based on the example of human skeleton video-based action recognition task. While \textbf{existing GNNs} can only process graphs whose vertices carrying \textit{\textbf{vectors or single values}}, they require to reduce each graph-style human skeleton frame as a vector. Our approach can directly include every graph-style frame in a GIG vertex in the proposed GIG sample, which is then processed by our GIG network that can process GIG samples whose vertices carrying \textit{\textbf{more comprehensive graphs}}.}
\label{fig:comparison_eg_gig}
\end{figure}

To extract task-specific patterns from graph vertices and their connections (i.e., edges) for analysis, most early GNNs \cite{perona1990scale,battaglia2016interaction,marcheggiani2017encoding,velivckovicdeep} were built upon anisotropic operations, which updates vertices based on their first-order vertex neighbours, while others \cite{scarselli2008graph,bruna2014spectral,defferrard2016convolutional,sukhbaatar2016learning,kipfsemi,velivckovicdeep} extended Convolution Neural Networks (CNNs) to the graph domain, allowing each vertex to exchange messages with its higher-order neighbours. Recently, more advanced GNNs are proposed to process graphs that have not only different typologies (heterogeneous graphs), e.g., Heterogeneous Graph Neural Network \cite{zhang2019heterogeneous}, Heterogeneous Graph Attention Network \cite{wang2019heterogeneous}, etc., but also multi-dimensional edge features (e.g., EGNNs \cite{gong2019exploiting}, ME-GCN \cite{wang2022me}, and MDE-GNN \cite{xiong2021multi}, etc.), aiming for more comprehensive and effective vertex and edge feature extraction/updating. Dwivedi et al. \cite{dwivedi2020benchmarking} further integrated the advanced heterogeneous graph and multi-dimensional edge feature processing mechanisms into several widely-used GNNs (e.g., Graph Attention Network (GAT) \cite{velivckovicgraph}, Gated Graph ConvNets (GatedGCN) \cite{2017arXiv171107553B}, etc.).

While there have been significant advancements in vertex and edge updating mechanisms for recent GNNs, to the best of our knowledge, no previous work has attempted to enhance another crucial aspect of GNNs, i.e., their vertex representing capability, which enables these GNNs to process more structured vertex representations (i.e., graphs). More specifically, since existing GNNs can only handle vertices represented by single values or vectors of a fixed dimension, previously proposed GNN-based real-world data analysis approaches largely depend on a data-specific pre-processing step that transform the target real-world data sample (e.g., video) as a graph, where each object/component (e.g., a video frame \cite{yan2018spatial}) of the data sample is usually encoded as a graph vertex represented by either a vector or a single value. This restricts the capability of GNNs to handle more complex scenarios, such as those where multiple complex objects are presented (e.g., a video consisting of multiple human face or skeleton frames), i.e., when representing each human face or skeleton, a graph may be a more powerful representation than a vector/value \cite{song2022gratis,luo2022learning,shi2019skeleton} (\textbf{Motivation 1}). Additionally, recent studies found that the relationships among graph samples in the same dataset can provide crucial cues to further enhance their analysis performance (i.e., enhancing performances for generic graph analysis tasks) \cite{zhong2021graph,pan2021multi}, where contrastive learning strategies have been frequently extended to model such relationships \cite{ji2023rethinking}. Such strategies usually utilize loss functions to maximise the similarity among graphs/vertices belonging to the same category while minimising the similarity among graphs belonging to the different categories, which are limited to model a specific relationship among graphs based on the manually-defined loss function. As a result, a more effective, comprehensive and task-specific strategy for modelling the relationship among graphs may further improve GNNs' performances in generic graph analysis tasks (\textbf{Motivation 2}).

In this paper, we propose a novel \textbf{Graph in Graph (GIG) Neural Network} which can process a more comprehensive graph-style data structure (called \textbf{GIG sample}) whose vertices can carry more powerful representations (i.e., graphs). The proposed GIG network is flexible to incorporate advanced vertex and edge updating functions from various existing GNNs, making it capable of handling GIG samples carrying multiple heterogeneous graphs and graphs containing multi-dimensional edge features. The comparison between the proposed GIG network and existing GNNs are visualised in Fig. \ref{fig:comparison_eg_gig}. Our GIG network starts with a \textbf{GIG sample generation (GSG) layer} which encodes the input data sample (a batch of graphs or a non-graph data sample whose components can be effectively represented by multiple graphs (called multi-graph data sample in this paper)) as a GIG sample consisting of multiple GIG vertices, where each vertex contains a graph. Then, a set of \textbf{GIG hidden layers} are stacked, where each is composed of two modules: a GIG vertex updating (GVU) module that independently updates the graph in each GIG vertex based on its internal information, and a global-level GIG sample updating (GGU) module that updates the graph in each GIG vertex based on its relationships with graphs in other GIG vertices. Finally, a \textbf{GIG output layer} is attached to produce the updated local and global context-aware GIG latent representation.
The full pipeline of our GIG network is also illustrated in Fig. \ref{fig: pipeline}. The main contributions and novelties of this paper are summarised as follows:
\begin{itemize}

\item The proposed GIG Network is a large improvement over existing GNNs, as it can process graph-style data (i.e., GIG samples) whose vertices can carry more structured representations (i.e., graphs). This diverges from existing GNNs \cite{velivckovicdeep,2017arXiv171107553B,zhang2019heterogeneous,wang2019heterogeneous,dwivedi2020benchmarking} that can only handle graphs with vertices containing vectors or single values (\textbf{Novelty 1}).

\item We propose a novel and efficient two-stage updating strategy to define the GIG hidden layer, where GVU and GGU modules can repeatedly and explicitly learn both internal cues within the graph contained in each GIG vertex as well as the relationship cues between graphs in different GIG vertices in an end-to-end manner. This enables our GIG network to process complex scenarios where each data sample can be effectively described by multiple graphs (\textbf{Novelty 2}).

\item Our GIG network is suitable for analyzing real-world multi-graph data (i.e., a data sample that can be effectively represented by multiple graphs), as well as a batch of graph samples for generic graph analysis tasks, where each input graph is contained in a GIG vertex. This differs from existing sub-graph GNNs \cite{alsentzer2020subgraph,sun2021sugar,meng2018subgraph,zhang2019quantum,wang2022glass,chen2022structure} which divide each input graph into multiple sub-graphs. The results show that our GIG network achieved state-of-the-art performances on 13 out of 14 datasets, highlighting its potential for a broad range of graph analysis tasks.

\end{itemize}

\section{Related Work}\label{sec2}
\subsection{Graph Neural Networks}
\label{subsec:related-GNN}
Graph Neural Networks (GNNs) are effective models to process graph-structured data, which models the intricate patterns and relationships inherent within graphs. Existing GNNs can be categorized as Message Passing GNNs (MPNNs) and Weisfeiler Lehman GNNs \cite{dwivedi2020benchmarking}. Specifically, MPNNs usually aggregate feature information from each vertex's nearest neighbours to update its state, potentially followed by a nonlinear transformation. For example, the Graph Convolution Network (GCN) \cite{kipf2016semi} is built on a straightforward message-passing mechanism, which aggregates information from each vertex's neighbouring vertices to update it. GraphSAGE \cite{hamilton2017inductive} extends such neighbour aggregation mechanism by learning a function for inductive feature aggregation from each vertex's local neighbouring vertices. Graph Attention Network (GAT) \cite{velivckovic2017graph} introduces an attention mechanism to weigh neighbours' contributions dynamically. Simple Graph Convolution (SGC) \cite{wu2019simplifying} simplifies the attention calculation by condensing consecutive convolutional layers into a single linear transformation, aiming to reduce computational demands. Jumping Knowledge Networks (JKN) \cite{xu2018representation} offer flexibility in aggregation by allowing features from various sizes of neighbour to be combined, to address the commonly faced over-smoothing issue in MPNNs. However, these MPNNs usually suffer from scalability challenges and might struggle with capturing global graph properties as they are focusing on local information aggregation.

Weisfeiler-Lehman GNNs (WLGNNs), on the other hand, push the boundaries of graph representation learning by capturing higher-order structural information. A typical example is the Graph Isomorphism Network (GIN) \cite{xu2018powerful} which maximizes the representational limit to match the Weisfeiler-Lehman isomorphism test, and thus enables the distinction of complex graph structures. Besides, K-GNN \cite{morris2019weisfeiler} expands maximization mechanism by generalizing the aggregation mechanism to higher-order vertex tuples, which can encapsulate more detailed structural information. DGCNN \cite{wang2019dynamic} is proposed for point clouds analysis, which adapts the Weisfeiler-Lehman (WL) principle to sort the vertex features in a way that corresponds to the order implied by the WL test. PAN \cite{ma2021path} enhances the learning capacity of GNNs by augmenting graph structures with paths that encapsulate higher-order connectivity patterns, which is similar to the WL kernel's attempt. This allows it to capture more complex graph structures beyond the nearest neighbours as the way employed in the GCN. Despite the advanced capabilities in capturing detailed and nuanced graph features, WLGNNs generally face the challenges including the increased computational demands as well as the higher probability to trigger overfitting during their training. Such drawbacks may make them harder to be implemented and optimized compared to their MPNNs counterparts.

\subsection{Graph vertex representation learning}
\label{subsec:related-vertex}
To utilise GNNs for real-world data analysis, a large number of graph representation learning strategies have been proposed to represent various types of data as graphs, where the key component/objects in the target data are usually represented as vertices in the graph. In the realm of graph-based image analysis, each image region or region containing an object is typically treated as a vertex feature vector. For example,  Li et al. \cite{li2017scene} and Han et al. \cite{han2022vision} propose scene graph generation approaches, where the image patch of each object in the scene image is encoded as a vertex vector in the corresponding scene graph, which are then processed by a GNN to model each scene's layout and its object interactions. Wang et al. \cite{Wang_2018_ECCV} introduce an graph-based video action recognition approach which represents each video as a graph. Specifically, each video frame (i.e., an image) is represented as a vertex feature vector in the graph. These vertices are then connected based on the temporal proximity of the corresponding frames and the similarity of their contents, to model their temporal relationships. Similar strategies have been proposed by Yan et al. \cite{yan2018spatial} and Chen et al. \cite{chen2021channel}, which also represent each video frame as a vertex vector in the graph. While spatial cues and relationships between different objects/components contained in each image patch or each video frame may be crucial for the downstream analysis, representing these cues within a vector may lead to significant information loss or distortion, and thus limit GNNs' analysis performances.

Graphs also have been used to represent audio data as valuable relationship exists among audio segments. In the work by Purwins et al. \cite{8678825}, a graph-based representation for music recommendation systems was proposed. Features of audio segments are extracted using signal processing techniques and then encoded into vectors to generate vertices. Similarly, An et al. \cite{shirian2022self}, Zhang et al. \cite{zhang2019few} and Singh \cite{singh2024atgnn} also generated vertices through directly encoding each audio segment as a vertex feature vector. While each audio segment typically possesses the abundant types of features extracted from different perspectives \cite{abbas2024artificial,kurzinger2020lightweight}, e.g., the time domain, the frequency domain and the Mel-Spectrogram etc, compressing all types of features within a vector could cause GNNs failing to well explore relationships among features extracted from different perspectives during the propagation, and thus limit their learning capability. Besides, graphs can also represent documents. For instance, Huang et al. \cite{huang2023text} introduced a graph-based text summarization method, where each sentence is encoded as a vertex feature vector. However, structural relationships (e.g., semantic characteristics, emotional status, or logical coherence) among words in each sentence could not be well represented within a vector, i.e., important contextual cues would be losing before the formal learning stage starts. 

\section{Preliminaries}
Given a graph sample $\mathcal{G}(V,E)$, where $V$ is a set of vertices with $v_n \in V$ and $E$ is a set of edges connecting adjacent vertices, existing message-passing GNN (MPNN) layers usually output an updated graph $\mathcal{\hat{G}}(\hat{V},\hat{E})$ as:
\begin{equation}
   \mathcal{\hat{G}}(\hat{V},\hat{E}) = f(\mathcal{G}(V,E), A),
\end{equation}
where $A$ is an adjacent matrix describing the topology of $\mathcal{G}(V,E)$. Each edge connecting vertex $v_n$ to $v_m$ is represented by $e_{n,m} = (v_n,v_m) \in E$. The generic function for updating each edge feature $e_{n,m}$ can be formulated as:
\begin{equation}
\begin{split}
    \hat{e}_{n,m} = 
    \begin{cases}
    f_e(e_{n,m}, v_n, v_m) & \text{Updating edge} \\
    e_{n,m} &  \text{Otherwise}
    \end{cases}
\end{split}
\label{eq:edge_basic}
\end{equation}
where $f_e$ is a differentiable edge feature updating function fed with the initial edge feature $e_{n,m}$ and corresponding vertex features $v_n$ and $v_m$. By updating all edges, the updated edge feature set $\hat{E}$ is obtained. Here, some GNNs (e.g., vanilla GCN \cite{kipf2016semi}, AdaGPR \cite{wimalawarne2023layer}, GATv2 \cite{brody2021attentive}) do not update edge feature, and thus the input and output graphs share same edge features. Then, each vertex feature $v_n$ is updated as:
\begin{equation}
\begin{split}
   \hat{v}_n &= f_v(v_n,\mathbf{m}_{\mathcal{N}(v_n)}), \\
   \mathbf{m}_{\mathcal{N}(v_n)} &= \text{Agg}(g_v(v_m,\hat{e}_{n,m}) \vert v_m \in \mathcal{N}(v_n))
\end{split}
\label{eq:vertex_basic}
\end{equation}
where $f_v$ is a differentiable vertex updating function that takes $v_n$ and its adjacent vertices $\mathcal{N}(v_n)$ as the input, while the operation Agg involves aggregating the information passed from adjacent vertices in a graph. Here, each adjacent vertex $v_m$ affects the target vertex $v_n$ via the edge that connects them, represented as $e_{m,n}$, and the the impact is also determined by the function $g_v$.

\begin{figure*}[t]
\begin{center}
\includegraphics[width=0.96\textwidth]{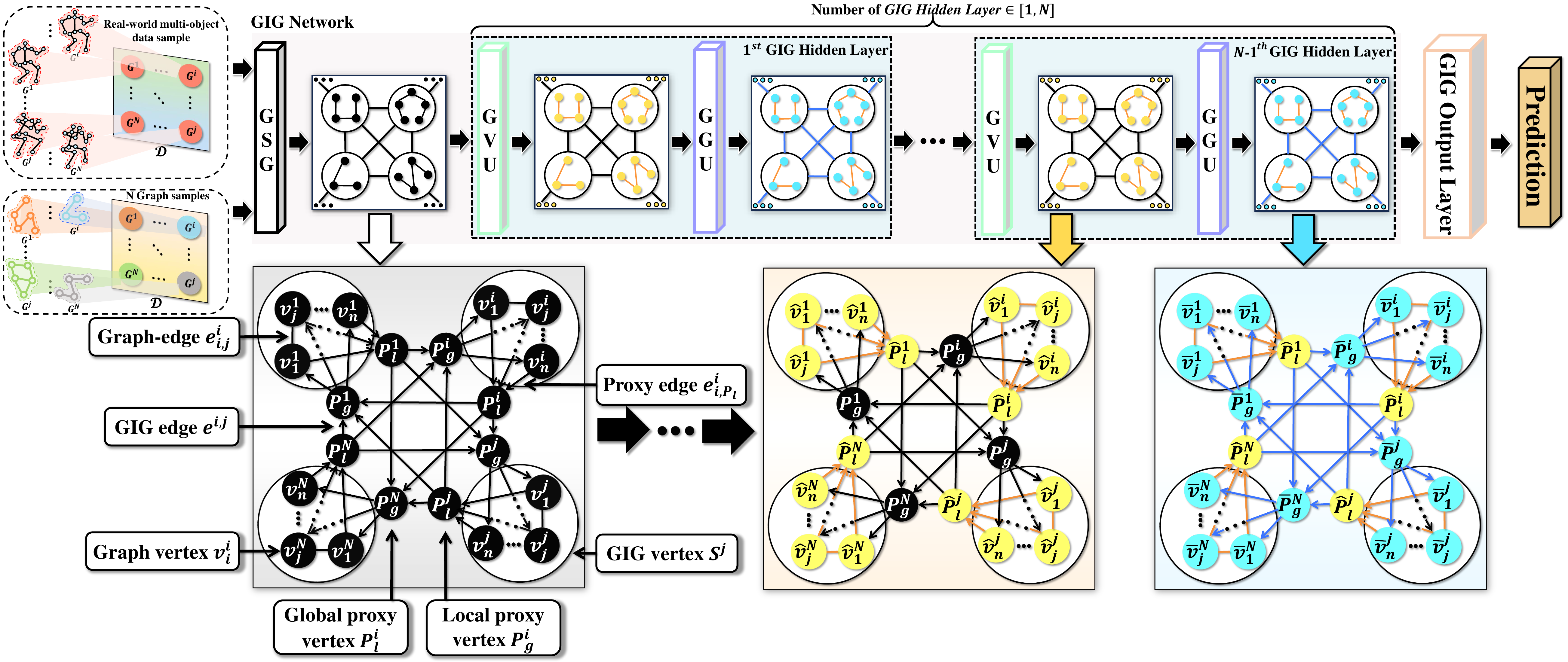}
\end{center}
   \caption{Illustration of the propagation of \textbf{the GIG network}. Given a real-world multi-graph sample or a set of graph samples as the input, the GSG layer first transforms the input as a \textbf{GIG sample}, where each graph/graph sample is represented by a GIG vertex and each GIG vertex is represented by a proxy vertex. Then, a set of GIG layers are stacked, where each consists of (i) a GVU module which individually updates each GIG vertex (i.e., its graph edges, graph vertices, directed proxy edges starting from graph vertices, and local proxy vertex) based on its internal cues; and (ii) a GGU module which further updates each GIG vertex based on its relationship with other GIG vertices  (i.e., updating global-edges, global proxy vertices, directed proxy edges starting from global proxy vertices, and graph vertices).}
\label{fig: pipeline}
\end{figure*}

\section{Graph in Graph Neural Network}

This section presents our novel GIG network. As demonstrated in Fig. \ref{fig: pipeline} and Algorithm \ref{Algorithm_GIG}, it starts with a \textbf{GIG Sample Generation (GSG)} layer (Sec. \ref{subsec: GIG data}), which transforms an arbitrary input data $\mathcal{D}$ to a GIG sample $\mathcal{G}(\mathcal{S}, P_l, P_g, E_P, E_G)$. In an obtained GIG sample, each key component/object of the input data is summarised in a graph contained in a GIG vertex $\mathcal{S}^{i}(V_S^i, E_S^i) \in \mathcal{S}$. Besides, this module also defines a local proxy vertex $P^{i}_l \in P_l$ (represented as a vector or a single value) to summarise the graph contained in each GIG vertex, as well as a global proxy vertex $P^{i}_g \in P_g$ aiming at collecting global and contextual information from graphs contained in other GIG vertices. To facilitate message exchanging among GIG vertices, the GSG layer initialises a set of undirected GIG edges $e^{i,j} \in E_G$ to connect each pair of GIG vertices via their local and global proxy vertices. Then, a set of \textbf{GIG hidden layers} are stacked to process the GIG sample output from the GSG layer, with each composed of a GIG vertex level updating (GVU) module \textbf{(Sec. \ref{subsec:GIG vertex updating})} and a global-level GIG sample updating (GGU) module  \textbf{(Sec. \ref{subsec:global-graph updating})}. The GVU module individually updates the graph contained in each GIG vertex by leveraging the internal (local) information within the graph, while the GGU module \textbf{(Sec. \ref{subsec:global-graph updating})} updates each GIG vertex based on its relationship with other GIG vertices, thereby updating GIG edges connecting GIG vertices. Finally, a \textbf{GIG output layer} is attached at the top of the GIG network to produce the final prediction.

\begin{algorithm}[t]
\begin{algorithmic}

\State \textbf{Input:} A original multi-graph data sample or a set of graph samples $\mathcal{D} = \{\mathcal{S}^1, \cdots, \mathcal{S}^I \}$.

\State \textbf{Output:} An updated GIG sample consisting of $I$ GIG vertices: $\overline{\mathcal{G}}^L(\overline{S}_G(\overline{V}_S, \overline{E}_S), \overline{P_g}, \overline{E}_P, \overline{E}_G)$

\State \textbf{Model:} GIG network equipped with a GSG layer, $L$ hidden GIG layers, and an output GIG layer.
\State

\State -----------------------------------------------------------------

\State \textbf{The propagation process of the GIG network:}

\State \textbf{GSG layer:} Producing a GIG sample from the input: $\mathcal{G}(S_G(V_S, E_S), P_l, P_g, E_P, E_G) \leftarrow \textbf{GSG}(\mathcal{D})$

\Statex \textbf{GIG hidden layers:} 
\For{l = 1:$L$}

\State \textbf{GVU module:} Locally and individually \qquad updating the graph edges $e^i_{n,m} \in E_S^i$, graph vertices $v^i_{n} \in V_S^i$, proxy edges $e^i_{n,P} \in E_P^i$ and local proxy vertices $P^i_l \in P_l$ in each GIG vertex $S^i$ as: $\hat{\mathcal{G}^l}(\hat{S}_G(\hat{V}_S, \hat{E}_S), \hat{P}_l, P_g, \hat{E}_P, E_G) \leftarrow \textbf{GVU}(\mathcal{G}^{l-1}(S_G (V_S, E_S), P_l, P_g, E_P, E_G))$

\State \textbf{GGU module:} Globally updating GIG edges $e^{i,j} \in E_G$ and global proxy vertices $P^i_g \in P_g$, and then pass global contexts to each GIG vertices by updating its proxy edges $e^i_{P,n} \in E_P^i$ and graph vertices $\hat{v}^i_{n} \in \hat{V}_S^i$ as: $\overline{\mathcal{G}}^l(\overline{S}_G(\overline{V}_S, \hat{E}_S), \hat{P}_l, \overline{P}_g, \overline{E}_P, \overline{E}_G) \leftarrow \textbf{GGU}(\hat{\mathcal{G}}^{l}(\hat{S}_G(\hat{V}_S, \hat{E}_S),\hat{P}_l, P_g, \hat{E}_P, E_G))$

\EndFor

\State \textbf{GIG output layer:} First applying the GVU module to individually update each GIG vertex, and then the GGU$^+$ is employed to not only updates GIG edges, global proxy vertices and graph vertices in each GIG vertex but also additionally updates graph edges as: $\overline{\mathcal{G}}^{L}(\overline{S}_G(\overline{V}_S, \overline{E}_S), \hat{P}_l, \overline{P}_g, \overline{E}_P, \overline{E}_G) \leftarrow \textbf{GGU}^+(\hat{\mathcal{G}}^L(\hat{S}_G(\hat{V}_S, \hat{E}_S), \hat{P}_l, P_g, \hat{E}_P, E_G))$

\end{algorithmic}
\caption{Pseudocode of the GIG network propagation}
\label{Algorithm_GIG}
\end{algorithm}


\subsection{GIG sample generation (GSG) layer}
\label{subsec: GIG data}

\textbf{GIG sample definition.} We first introduce a novel graph-style structure called Graph in Graph (GIG) sample (denoted as $\mathcal{G}$), which is made up of a set of GIG vertices $\mathcal{S}$ and GIG edges $E_G$. Specifically, each GIG vertex $\mathcal{S}^{i}$ in $\mathcal{G}$ contains a graph $\mathcal{S}^{i}(V_S^i, E_S^i)$ that is made up of $N_i$ vertex vectors $V_S^i = \{v_{1}^i, v_{2}^i, \cdots, v_{N_i}^i\}$ (called graph vertices in this paper), and a set of edges (called graph edges in this paper), where each graph edge $e_{n,m}^{i} \in E_S^i$ connects a pair of graph vertices $v_{n}^i$ and $v_{m}^i$ in $\mathcal{S}^{i}$ (illustrated in Fig. \ref{fig: pipeline}). To facilitate efficient communication between graphs contained in different GIG vertices, the GIG sample also defines a local proxy vertex $P^i_l \in P_l$ and a global proxy vertex $P^{i}_g \in P_g$ for each GIG vertex $\mathcal{S}^{i}$, where $P^i_l$ integrates local features from the graph contained in $\mathcal{S}^{i}$, while the global proxy vertex $P^{i}_g$ receives contextual cues from other GIG vertices, aiming to further pass them to update graph vertices of the $\mathcal{S}^{i}$. As a result, each local proxy vertex $P^i_l$ / global proxy vertex $P^i_g$ is linked to a set of graph vertices contained in its GIG vertex $\mathcal{S}^{i}$ (e.g., directed edge $e^i_{n,P_l}$ from the graph vertex $v^i_n$ to local proxy vertex $P^i_l$ / directed edge $e^i_{P_g,n}$ from global proxy vertex $P^i_g$ to the graph vertex $v^i_n$). Meanwhile, each pair of connected GIG vertices $S^i$ and $S^j$ exchange messages via a pair of directed global-edges $e_{i,j} \in E_G$ and $e_{j,i} \in E_G$, where $e_{i,j}$ starts from the local proxy vertex $P^i_l$ of $S^i$ to the global proxy vertex $P^j_g$ of $S^j$ passes the cues of $S^i$ to the GIG vertex $S^j$. Thus, a GIG sample can be represented as $\mathcal{G}(S_G (V_S, E_S), P_l, P_g, E_P, E_G)$. The definitions of employed notations are listed in Table \ref{tab:notation}.


\textbf{GSG layer.} The GSG layer transforms an arbitrary input data sample $\mathcal{D}$ (i.e., real-world multi-graph data or a batch of graph samples) to a GIG sample accordingly, where each GIG vertex contains either a graph representation representing an object/component of the input multi-graph data or a graph sample of the input graph batch. For example, for human skeleton video-based action recognition tasks, each human skeleton frame will be summarised as a graph and contained in a GIG vertex. Consequently, the GIG sample is constructed on the basis of multiple frames, with each GIG vertex representing a skeleton frame. For generic graph analysis tasks, a batch of graph samples will be combined as a GIG sample, with each GIG vertex containing a graph sample. After defining all GIG vertices, the GSG layer initialises each local proxy vertex $P^i_l$ by averaging all graph vertices in its corresponding GIG vertex $\mathcal{S}^i$, and each global proxy vertex $P^i_g$ as a zero vector. Then, the GSG defines a set of directed proxy edges to connect a set of graph vertices with its corresponding local proxy vertex $P^i_l$ and global proxy vertex $P^i_g$, and initialises them as zero vectors. For each GIG vertex, its proxy edges are connected: (i) from a set of its graph vertices that are least similar to its local proxy vertex; and (ii) from its global proxy vertex to a set of its graph vertices that are most similar to the local proxy vertex, where cosine similarity is empirically used in this paper (the influence of different similarity measurements are evaluated in Supplementary Material). The experimental analysis for various connection strategies are reported in Table \ref{tab:number of proxy edges} and \ref{tab:synthesized settings}, indicating that our approach is robust to not only different numbers of graph vertices connected to local/global proxy vertices but also different local proxy vertex initialisation strategies. In comparison to directly connect graph vertices contained in different GIG vertices, our proxy vertex and GIG edge-based strategy avoids the GIG sample to have extremely large number of edges to construct associations between GIG vertices. In summary, the GSG layer encodes the input data $\mathcal{D}$ as an GIG sample $\mathcal{G}(S_G(V_S, E_S), \mathbf{P}_l, \mathbf{P}_g, E_P, E_G)$ (denoted as $\mathcal{G}$ in the following sections).

\begin{table}[t]
    \caption{Notations for describing the initial GIG sample.}
    \label{tab:notation}
    \centering
    \begin{tabular}{ll}
    \hline
         \textbf{Notations}& \textbf{Descriptions} \\
         \hline
        \textbf{GIG sample}  & Graph-style data whose vertices  \\
                                    & can directly include graphs \\
        \textbf{GIG network/layer}  & Graph Neural Network (layer) \\ 
                                    & that can process GIG samples \\ \hline
         $\mathcal{G}(S_G, \mathbf{P}_l, \mathbf{P}_g, E_P, E_G)$  & GIG sample \\
         $S_G(V_S, E_S)$ & the vertex/GIG vertex set of $\mathcal{G}$ \\
         $\mathcal{S}^i(V^i_S, E^i_S)$ & a vertex/GIG vertex in $\mathcal{G}$\\  
         $V^i_S$ & the graph vertex set in GIG \\
                 & vertex $S^i$ \\
         $v^i_n$& a graph vertex in $V^i_S$\\ 
         $E^i_S$ & the graph edge set in GIG \\
                 & vertex $S^i$ \\
         $e_{n,m}^i$ & a graph edge in $E^i_S$ connecting \\
                     & $v^i_n$ and $v^i_m$ \\
         $\mathbf{P}_l$ & the local proxy vertex set in $\mathcal{G}$ \\
         $\mathbf{P}_l^i$& the local proxy vertex \\ 
                         & representing $\mathcal{S}^i$ \\
         $\mathbf{P}_g$ & the global proxy vertex set in $\mathcal{G}$ \\
         $\mathbf{P}_g^i$& the global proxy vertex of $\mathcal{S}^i$ \\
                       & collecting information from other  \\ 
                       & GIG vertices $\mathcal{S}^j$ \\
         $E_P^i$ & the set of directed proxy edges  \\
                 & for $\mathcal{S}^i$ \\
         $e_{n,P_l}^i$ / $e_{P_g,n}^i$ & a pair directed proxy edges   \\
                       & connecting $v^i_n$ and $\mathbf{P}^i_l$ or $\mathbf{P}^i_g$ and \\
                       & $v^i_n$ \\
         $E_G$ & the set of GIG edges in $\mathcal{G}$\\
         $e_{i,j}$ & a GIG edge in $E_G$ connecting \\  
                   & $\mathbf{P}^i_l$ and $\mathbf{P}^j_g$ \\
    \hline  
    \end{tabular}

\end{table}

\subsection{GIG hidden layer}

In the proposed GIG network, a set of GIG hidden layers are stacked after the GSG layer to learn downstream task-related features from the obtained GIG sample. Specifically, each GIG hidden layer consists of a \textbf{GVU module} which individually updates the graph contained in each GIG vertex based on its internal/local information, as well as a \textbf{GGU module} which enables the graphs contained in different GIG vertices to exchange their task-related cues. This allows each GIG hidden layer jointly models local information contained in each GIG vertex and relationships among GIG vertices, facilitating effective and global context-aware message exchanging during reasoning.

\begin{figure*}[t]
\begin{center}
\includegraphics[width=0.96\textwidth]{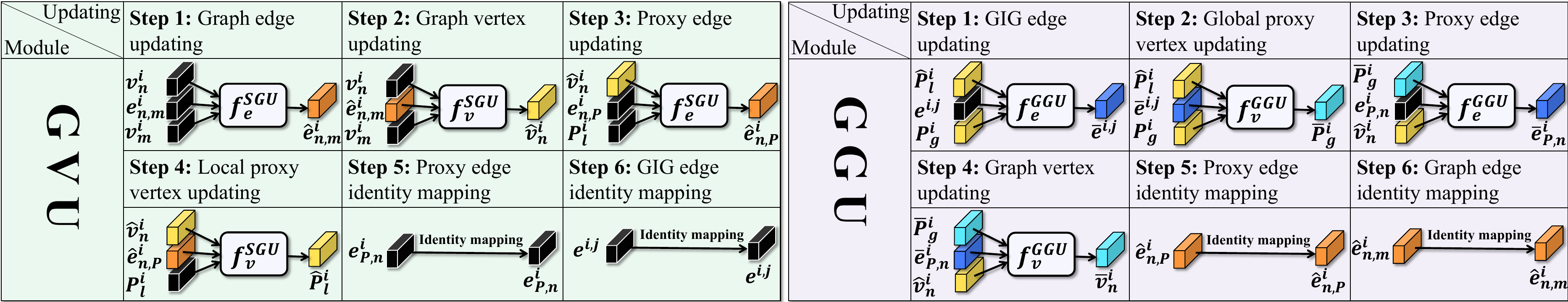}
\end{center}
   \caption{Illustration of the propagation of \textbf{the GVU and GGU module}.}
\label{fig: gvu_ggu_workflow}
\end{figure*}

\subsubsection{GIG Vertex Updating (GVU) module} 
\label{subsec:GIG vertex updating}

Let the input to the GVU module be a GIG sample $\mathcal{G}$, it first independently updates the graph contained in each GIG vertex $\mathcal{S}^i$, including its graph edges and graph vertices, and then passes such updated local cues to the corresponding local proxy vertex $P^i_l$ via the updated directed proxy edges starting from a set of graph vertices $v^i_n$ to $P^i_l$. 

\textbf{Locally updating graph edges and graph vertices:} This module first updates each \textbf{graph edge feature $e_{n,m}^i$} in a GIG vertex $\mathcal{S}^i$ by considering: (i) the $e_{n,m}^i$ itself; and (ii) the corresponding graph vertices $v_{n}^i$ and $v_{m}^i$ as:
\begin{equation}
\hat{e}_{n,m}^i = f^\text{GVU}_e(e_{n,m}^i, v_{n}^i, v_{m}^i)
\label{GIG vertex:edge}
\end{equation}
where $f^\text{GVU}_e$ represents the graph edge feature updating function employed by the GVU. It then updates each \textbf{graph vertex feature $v_{n}^i$} by considering: (i) the graph vertex feature $v_{n}^i$; (ii) $v_{n}^i$'s adjacent graph vertex features $v_{m}^i \in \mathcal{N}_{\text{local}}(v_n^i)$ in the corresponding GIG vertex $\mathcal{S}^i$; and (iii) the updated graph edge $\hat{e}_{n,m}^i$ that connects $v_{n}^i$ and its adjacent graph vertices $\mathcal{N}_{\text{local}}(v_n^i)$. This can be formulated as:
\begin{equation}
    \hat{v}_{n}^{i} = f^\text{GVU}_v( v_{n}^{i}, \text{Agg}(g_v(\hat{e}_{n,m}^{i}, v_{m}^{i}) \vert  v_m^{i} \in \mathcal{N}_{\text{local}}(v_n^i)))
\label{GIG vertex:node}
\end{equation}
where $f^\text{GVU}_v$ is a differentiable vertex feature updating function in the GVU; $g_v$ is a differentiable message passing function that passes the impact of each vertex feature to its neighbours; and $\text{Agg}$ denotes an aggregation operation.

\textbf{Locally updating proxy edges and local proxy vertices:} Once all graph edges and graph vertices are updated, the GVU also updates each \textbf{directed proxy edge $e_{n,P_l}^i$} starting from each graph vertex $v_{n}^i$ to its local proxy vertex $P^i_l$ by considering: (i) the proxy edge $e_{n,P_l}^i$; (ii) the corresponding updated graph vertices $\hat{v}_{n}^i$; and (iii) the local proxy vertex $P^i_l$, which can be denoted as:
\begin{equation}
\hat{e}_{n,P_l}^i = f^\text{GVU}_e(e_{n,P_l}^i, \hat{v}_{n}^i, P^i_l)
\label{GIG vertex:proxy edge}
\end{equation}
where the $f^\text{GVU}_e$ is re-employed as the proxy edge feature updating function. Finally, the \textbf{local proxy vertex $P^i_l$} is updated by aggregating messages passed from selected graph vertices (which are least similar to the local proxy vertex) through the corresponding updated directed proxy edges $\hat{e}_{n,P_l}^i$ ($n = 1,2,\cdots, N_i$). This process can be formulated as:
\begin{equation}
    \hat{P}^{i}_l = f^\text{GVU}_v (P^{i}_l, \text{Agg}(g_v(\hat{e}_{n,P_l}^{i}, \hat{v}_n^i)))
\end{equation}
where $f^\text{GVU}_v$ is re-employed as the local proxy vertex updating function. This way, each local proxy vertex $\hat{P}^i_l$ summarises the locally updated graph information contained in its corresponding GIG vertex $\hat{\mathcal{S}}^i$. It should be emphasised that we follow previous GNNs \cite{simonovsky2017dynamic,bresson2017residual,cui2020edge,shang2018edge,schlichtkrull2018modeling} to first update edges (e.g., graph edges / proxy edges), and then updates the corresponding vertices (graph vertices / local proxy vertices / global proxy vertices). Also, we found that first updating graph edges/vertices and and then updating proxy edges/vertex lead to similar performances as simultaneously updating them. \\

\subsubsection{Global-level GIG Sample Updating (GGU) Module}
\label{subsec:global-graph updating}

After individually updating each GIG vertex, the GGU module aims to exchange messages among GIG vertices. It first passes locally updated cues from the local proxy vertex of each GIG vertex to the global proxy vertices of its adjacent GIG vertices, via the updated GIG edges. Then, the contextual information received in each global proxy vertex (i.e., containing cues passed from local proxy vertices of other GIG vertices) is further passed to its corresponding GIG vertex, enabling the contained graph to be updated based on contextual cues provided by graphs contained in other GIG vertices.

\textbf{GIG edge and global proxy vertex updating:} Given a directed GIG edge $e^{i,j} \in E_G$ starting from the locally updated local proxy vertices $\hat{P}^{i}_l$ of the GIG vertex $S^i$ to the global proxy vertex $P^{j}_g$ of the GIG vertex $S^j$, the GGU updates it via a differentiable GIG edge updating function $f_e^\text{GGU}$ as:
\begin{equation}
\overline{e}^{i,j} = f_e^\text{GGU}(e^{i,j}, \hat{P}^{i}_l, P^{j}_g)
\label{global-graph:edge}
\end{equation}
Then, the GGU \textbf{exchanges messages among GIG vertices by updating their global proxy vertices $\hat{P}^{i}_g$ as $\overline{P}^{i}_g$} based on: (i) the global proxy vertex feature $P^{i}_g$; (ii) the local proxy vertices $\hat{P}^{j}_l \in \mathcal{N}_\text{global}(\hat{P}^{i}_g)$ ($j = 1,2, \cdots, I$ and $j \neq i$) of its neighbouring GIG vertices, which summarises the locally updated graphs in their corresponding GIG vertices; and (iii) the corresponding globally updated directed GIG edges $\overline{e}^{i,j}$ ($j = 1,2, \cdots, I$ and $j \neq i$) as:
\begin{equation}
    \overline{P}^{i}_g = f_v^\text{GGU}( P^{i}_g, \text{Agg}(g_v(\overline{e}^{i,j}, \hat{P}^{j}_l) \vert \hat{P}^{j}_l \in \mathcal{N}_{\text{global}}(P^i_g)))
\label{global-graph:vertex}
\end{equation}
where $f_v^\text{GGU}$ denotes a differentiable global proxy vertex feature updating function in the GGU. This way, each updated global proxy vertex would contain messages received from other locally updated GIG vertices, i.e., the updated global proxy vertices are global context-aware.

\textbf{Globally updating proxy edges and graph vertices:} Finally, the GGU passes the global contextual information received in each globally updated global proxy vertex $\overline{P}^{i}_g$ to the graph in the corresponding GIG vertex $\mathcal{\hat{S}}^{i}$. This is achieved by first updating each \textbf{directed proxy edge feature $e_{P_g,n}^i$} starting from $\overline{P}^i_g$ to each of its connected graph vertex $\hat{v}_{n}^i$ by considering: (i) the $\overline{P}^{i}_g$ that contains messages received from other GIG vertices; (ii) the directed proxy edge $e_{P_g,n}^i$; and (iii) the locally updated graph vertex $\hat{v}_{n}^i$ as:
\begin{equation}
    \overline{e}_{P_g,n}^i = f^\text{GGU}_e(e_{P_g,n}^i, \overline{P}^i_g, \hat{v}_{n}^i)
\label{GIG vertex:representative_edge}
\end{equation}
where $f^\text{GGU}_e$ is employed for updating directed proxy edges starting from global proxy vertices.

Then, the GGU updates graph vertex features that are connected with its global proxy vertex $\overline{P}^{i}_g$ based on the global contextual information by considering: (i) each locally updated graph vertex $\hat{v}_{n}^i$ that is connected with the global proxy vertex $\overline{P}^{i}_g$; (ii) the globally updated global proxy vertex $\overline{P}^i_g$; and (iii) the corresponding globally updated proxy edge $\overline{e}_{P_g,n}^i$ as:
\begin{equation}
    \overline{v}_{n}^{i} = f^\text{GGU}_v( \hat{v}_{n}^{i},  \text{Agg}(g_v(\overline{e}_{P_g,n}^{i}, \overline{P}^{i}_g)))
\label{GIG vertex:ggu_node}
\end{equation}
where $f^\text{GGU}_v$ updates these graph vertices based on the global contextual information encoded in the corresponding global proxy vertex. Since a GIG network would contain multiple GIG hidden layers, the global contextual information encoded in graph vertices that are connected with their global proxy vertices at the $l_{th}$ hidden layer will be further utilized to update their related graph edges and the rest graph vertices by the GVU module at the ${l+1}_{th}$ hidden layer or the GIG output layer. Therefore, the GGU module of each GIG hidden layer does not need to globally update graph edges and rest graph vertices.

\subsection{GIG output layer}
\label{subsec:gig_output}

The GIG output layer has a similar architecture as the GIG hidden layer, which stacks a GVU module and a GGU$^+$ module. Specifically, its GVU module individually updating each GIG vertex (i.e., locally updating  graph edges and graph vertices of the graph) as well as its corresponding proxy edges and local proxy vertices based on the same functions defined by the GIG hidden layer. Then, the GGU$^+$ module follows the same manner as the GGU module defined in the GIG hidden layer to first globally update GIG edges, global proxy vertices, proxy edges starting from global proxy vertices and graph vertices. In addition, it further updates each locally updated graph edge $\hat{e}_{n,m}^i$ as $\overline{e}_{n,m}^i$ in each GIG vertex by considering: (i) the $\hat{e}_{n,m}^i$ itself; and (ii) the corresponding globally updated graph vertices $\overline{v}_{n}^i$ and $\overline{v}_{m}^i$ as:
\begin{equation}
\overline{e}_{n,m}^i = f_e^\text{GVU}(\hat{e}_{n,m}^i, \overline{v}_{n}^i, \overline{v}_{m}^i)
\label{output_layer:edge}
\end{equation}
where $f_e^\text{GVU}$ is re-employed. Subsequently, the graph vertices that are not connected with global proxy vertices are also updated based on these globally updated graph vertices $\overline{v}_{n}^i$ and $\overline{v}_{m}^i$ as well as other globally updated graph vertices (i.e., graph vertices connected with global proxy vertices) under the same rule of Eqa. \ref{GIG vertex:node}. This way, all graph vertices and graph edges of graphs contained in the final output GIG sample are locally and globally updated. Note that our GIG network allows for flexible customization of its edge and vertex updating functions from existing GNNs (e.g., GatedGCN \cite{2017arXiv171107553B} and GAT \cite{velivckovicgraph}) and thus it can jointly handle graphs with varying typologies and multi-dimensional edge graphs. To keep a simple form, this paper maintains a consistent function for all edges' updating (i.e., $f_e^\text{GVU}$ and $f_e^\text{GGU}$ have the same form), as well as the same vertex updating function for all types of vertices' updating (i.e., $f_v^\text{GVU}$ and $f_v^\text{GGU}$ are the same).

\subsection{Model complexity analysis}
\label{subsec:complexity analysis}

The GIG network consists of a GSG layer, a set of GIG hidden layers and a GIG output layer. Specifically, \textbf{the GSG layer} defines each local proxy vertex through averaging all graph vertex features of the corresponding GIG vertex. The averaging calculation for each GIG vertex consists of one summation operation and one division operation. The summation operation requires $O(n)$ and division operation requires $O(1)$. Then the time complexity for local proxy vertex generation procedure can be defined as $O(|\mathcal{G}| \times n)$. In this paper, the generation of proxy-edges are totally based on the cosine similarity between graph vertices and their corresponding local/global proxy vertex, which is defined as:
\begin{align}
\text{CosSim}(P_l^i, v_n^i) &= \frac{P_l^i \cdot v_n^i}{\|P_l^i\|\|v_n^i\|} \cr
                            &= \frac{\sum_{i=1}^{n} P_{l,i}^i v_{n,i}^i}{\sqrt{\sum_{i=1}^{n} {P_{l,i}^i}^2} \sqrt{\sum_{i=1}^{n} {v_{n,i}^i}^2}}    
\end{align}
where $P_l^i$ represents a local proxy vertex feature and $v_n^i$ represents one of its graph vertex feature, i.e.,  $P_{l,i}^i$ and $v_{n,i}^i$ denote the $i_\text{th}$ element in $P_l^i$ and $v_n^i$, respectively. Specifically, this calculation process involves three steps: (i) Dot Product calculation: The dot product \(P_l^i \cdot v_n^i\) necessitates \(n\) multiplication operations (one for each pair of corresponding elements from vectors \(P_l^i\) and \(v_n^i\)) and \(n-1\) addition operations to sum these products. Hence, the time complexity for computing the dot product is $O(n)$; (ii) Norm calculation: The norm of a vector, \(\|P_l^i\|\) or \(\|v_n^i\|\), requires \(n\) squaring operations, \(n-1\) addition operations to aggregate these squares, and a square root operation. Since both vectors undergo this process, the complexity remains \(O(n)\) for each norm calculation; and (iii) Division operation:  dividing the dot product by the product of the norms involves a single division operation, which is considered as \(O(1)\) in complexity. In summary, the overall time complexity for computing the cosine similarity between two \(n\)-dimensional vectors is $O(n)$. Continuously, the main operation in proxy edge generation is to calculate the cosine similarity between local proxy vertex and graph vertices in each GIG vertex. Since the numbers of graph vertices in different GIG vertices could be different, they can be represented as $V_S = \{|V_S^1|,|V_S^2|,|V_S^3|,...,|V_S^N|\}$. Consequently, the time complexity of proxy edge generation procedure can be defined as $O(|\mathcal{G}| \times max(V_S) \times n)$, where $|\mathcal{G}|$ represents the number of GIG vertices in the GIG sample. Then, the connections (GIG edges) between global proxy vertices and local proxy vertices from other GIG vertices are defined by computing self-cosine similarity on the matrix consisted of $|\mathcal{G}|$ local proxy vertex features (GSG module only works before training, while the global proxy vertex features are set as zero vectors here), resulting in the time complexity of $O(|\mathcal{G}|^2 \times n)$. Consequently, the overall time complexity of the GSG module is $O(|\mathcal{G}| \times max(V_S) \times n+|\mathcal{G}|^2 \times n)$. \textit{The GSG module will only be called once before the starting of training to generate the GIG sample.} Then, each \textbf{GIG hidden layer} is made up of a GVU and a GGU module, whose exact time complexity is determined by the edge/vertex updating algorithms chosen here. Assuming that the total time complexity of the GVU and GGU are $O(f_1)$ and $O(f_2)$, respectively, the time complexity of each GIG hidden layer can be defined as $O(f_1 + f_2)$. Finally, \textbf{GIG output layer} additionally updates graph vertices that are not connected with global proxy vertices as well as graph edges compared to GIG hidden layer, whose time complexity can be denoted as $O(f_3)$ and $O(f_4)$. As a result, the total time complexity of each GIG output layer can be defined as $O(f_1+f_2+f_3+f_4)$. The time complexity analysis of example GIG-GatedGCN and GIG-GAT, practical time consumption of our GIG network and the influence of the numbers of proxy edges and GIG edges on the GVU and GGU's complexity are provided in the Supplementary Material.

\section{Experiment}
\label{sec:experiments}

This section comprehensively evaluates our GIG on 14 different datasets (explained in Sec. \ref{subsec:dataset}), where the implementation details and evaluation metrics are detailed in Sec. \ref{subsec:implemntation}. We show that our GIG-GNNs which represent each vertex as a graph are superior to existing GNNs that represent each vertex as a vector/single value in Sec. \ref{subsec:SOTA}. Finally, Sec. \ref{subsec:Ablation} comprehensively evaluate various aspects of our GIG in terms of both effectiveness and efficiency.

\subsection{Datasets}
\label{subsec:dataset}

This paper evaluate our approach on 14 graph and non-graph datasets. Firstly, \textbf{12 benchmark graph datasets} are employed to evaluate the performance of the proposed GIG network on \textbf{generic graph analysis}, including (i) MNIST \cite{lecun1998gradient} and CIFAR10 \cite{krizhevsky2009learning} image classification datasets (i.e., their images are converted into graphs by a widely-used graph benchmark \cite{dwivedi2020benchmarking}),  three TU datasets \cite{zhao2018work, dobson2003distinguishing, gallicchio2020fast}, and the OGBG-PPA biological graph dataset (describing protein-protein associations) dataset for graph classification; (ii) ZINC \cite{irwin2012zinc}, ZINC-full \cite{irwin2012zinc} and AQSOL \cite{sorkun2019aqsoldb} molecular graph datasets for graph regression; (iii) PATTERN node-level graph pattern recognition \cite{dwivedi2020benchmarking} and CLUSTER semi-supervised graph clustering \cite{dwivedi2020benchmarking} datasets for node (vertex) classification; and (iv) TSP city route plan dataset \cite{joshi2022learning} for the link prediction. Then, our GIG is evaluated on \textbf{two non-graph real-world multi-graph datasets}, i.e., two subsets of the NTU RGB+D human skeleton video-based action recognition dataset which contains 56,880 skeleton action clips belonging to 60 classes (performed by 40 different participants). Here, each clip is recorded through three Microsoft Kinect v2 cameras at horizontal angles of $45^{\circ}$, $0^{\circ}$, $-45^{\circ}$, and contains human actions performed by one or two subjects. The employed two subsets are: (i) Cross-Subject (X-Sub) dataset, where clips of 20 subjects are used for training, and clips of the remaining 20 subjects are used for validation; and (ii) Cross-View (X-View) dataset, where two of the three camera-views are utilized for training, and the other is utilized for validation. 

\subsection{Implementation details}
\label{subsec:implemntation}

To demonstrate the effectiveness of our GIG network, we apply it to 14 datasets for 5 different tasks, where different vertex and edge updating algorithms inherited from GNNs are employed. Detailed training settings (e.g., hyper-parameter settings) are provided in the Supplementary Material.

\textbf{Experiments on graph datasets:} We individually apply the vertex/edge updating mechanisms of the widely-used GatedGCN (the default setting), GAT and GCN implemented by \cite{dwivedi2020benchmarking}, as well as the DeeperGCN \cite{li2020deepergcn} to construct our GIG network for generic graph analysis experiments. Here, a batch of input graph samples are combined as a GIG sample by the GSG layer, where each GIG vertex includes a graph sample. The initial feature of each local proxy vertex is obtained through averaging all of its graph vertex features, while the initial feature of each global proxy vertex/proxy edge is set as a zero vector. Then, we customize the edge/vertex updating functions of GVU and GGU modules based on the GatedGCN, GAT, GCN and DeeperGCN, respectively. During training,  we follow the same training settings (e.g., loss functions, hyper-parameters, and AdamW optimizer \cite{}) as the benchmarking \cite{dwivedi2020benchmarking} and DeeperGCN. The detailed evaluation metrics for these generic graph analysis tasks are: (i) graph classification results on MNIST \cite{lecun1998gradient}, TUs \cite{dwivedi2020benchmarking} and CIFAR10 \cite{krizhevsky2009learning} datasets are measured by classification accuracy; (ii) graph regression results on ZINC, ZINC-full \cite{irwin2012zinc} and AQSOL \cite{sorkun2019aqsoldb} datasets are measured by Mean Absolute Error (MAE); (iii) node (vertex) classification results on PATTERN \cite{dwivedi2020benchmarking} and CLUSTER \cite{dwivedi2020benchmarking} are measured by classification accuracy; (iv) link prediction results on TSP \cite{joshi2022learning} and OGBG-PPA \cite{hu2020open} are measured by F1-score and  classification accuracy, respectively.

\textbf{Experiments on human skeleton video-based action recognition datasets:} This paper additionally customize our GIG based on the edge/node updating algorithms of the state-of-the-art CTR-GCN \cite{chen2021channel} and widely-used ST-GCN \cite{yan2018spatial}, resulting in GIG-CTR-GCN and GIG-ST-GCN models for human skeleton video-based action recognition task. Here, each human skeleton video is encoded as a GIG sample by the GSG layer, where each skeleton frame is described as a graph contained in a GIG vertex while initial local/global proxy vertices and proxy edges are defined following the same rule described above. For fair comparsion, we follow the same training settings as the original ST-GCN and CTR-GCN, and the evaluation metrics is the Top-1 classification accuracy.

\begin{figure}[h!]
    \centering
    \subfigure[]{ 
        \includegraphics[width=.47\textwidth]{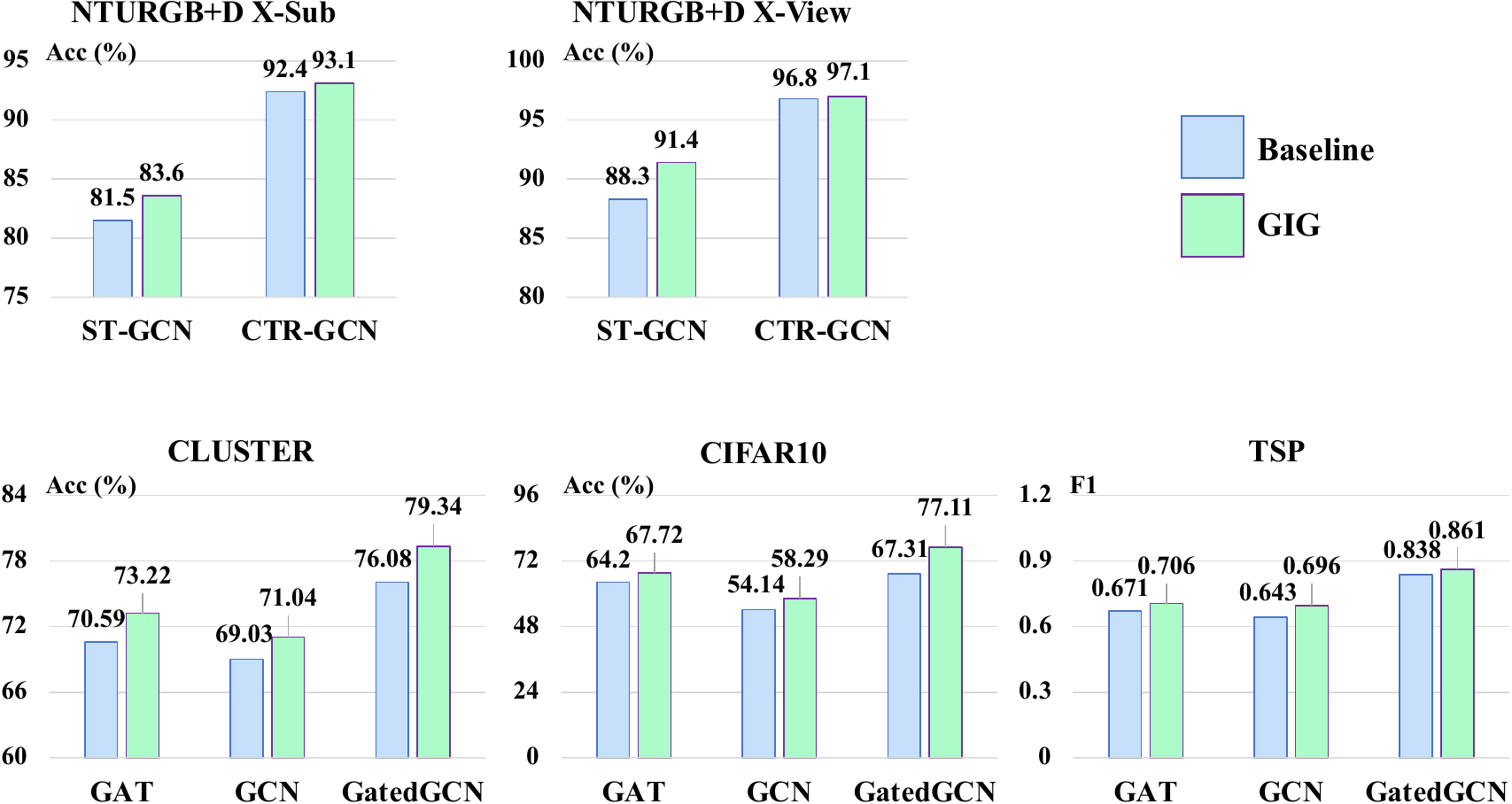} 
        \label{fig:flexibility}
    }
    \hfill 
    \subfigure[]{ 
        \includegraphics[width=.47\textwidth]{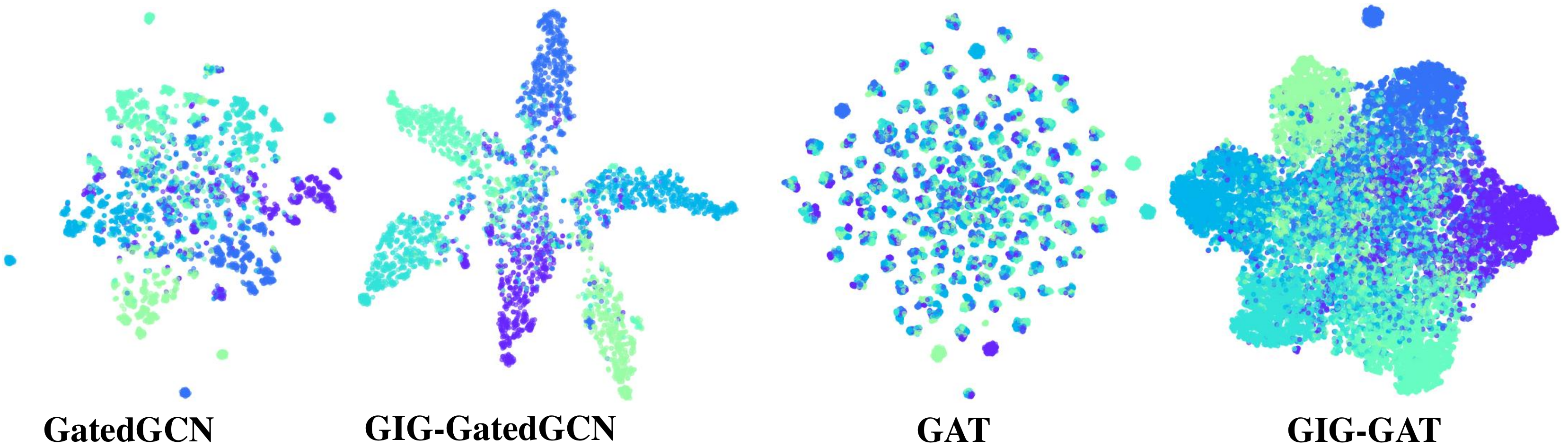} 
        \label{fig:visualization_tsne}
    }
    \caption{(a) Comparison between original GNNs and GIG-GNNs with the same vertex/edge updating functions;(b) t-SNE visualisation of features learned on CLUSTER dataset (a six classes vertex classification dataset), where each dot represents a vertex and each colour represents a category label.}
\end{figure}
\subsection{Comparison with state-of-the-art methods}
\label{subsec:SOTA}

This section compares our GIG network with existing approaches on various tasks. Fig. \ref{fig:flexibility} first validates that our GIG network is flexible to be customized by edge/vertex updating strategies defined by various existing GNNs. Importantly, \textit{with the same edge/vertex updating strategies, our graph vertex-based GIG consistently achieved performance gains over existing vector vertex-based GNNs for all evaluated tasks.} Besides, Fig. \ref{fig:visualization_tsne} demonstrates that representations learned by our GIG networks are more discriminative than baseline GNNs that have the same edge/vertex updating functions. This indicates that relationships among graph samples encoded by our GIG can provide complementary and task-specific cues for generic graph analysis, as well as representing and processing multiple objects (graphs) in multi-graph data samples.

\begin{table*}[htbp]
\caption{Experimental results on 11 benchmark datasets from \cite{dwivedi2020benchmarking}. Bolder results denote the best performance. Unfilled cells denote that these publications haven't evaluated their methods on the corresponding datasets.}
\label{tab:Benchmark results}
\centering
\small
\begin{tabular}{l|ccccc}
    \toprule
    \multirow{1}{*}{Task}& 
    \multicolumn{5}{c}{Graph classification}  \cr
    \hline
    Dataset                                         & MNIST                                   & CIFAR10           & ENZYMES            & DD                 &PROTEINS                                       \cr
    \multirow{.8}{*}{Method}& \multicolumn{5}{c}{Acc(\%) $\uparrow$}  \cr
    \hline
        GCN\cite{kipfsemi}                          & $90.12\pm0.15$                          & $54.14\pm0.39$    & $65.833\pm4.610$   &$72.758\pm4.083 $   &$76.098\pm2.406$                               \cr
        GIN\cite{xupowerful}                        & $96.49\pm0.25$                          & $55.26\pm1.53$    & $65.333\pm6.823$   &$71.910\pm3.873$    &$74.117\pm3.357$                               \cr
        $\delta$-2-GNN\cite{morris2020weisfeiler}   & -                                       & -                 & -                  &-                   &-                                              \cr
        $\delta$-2-LGNN\cite{morris2020weisfeiler}  & -                                       & -                 & -                  &-                   &-                                              \cr
        \hline
        GAT\cite{velivckovicgraph}                  & $95.54\pm0.21$                          & $64.22\pm0.46$    &$68.500\pm5.241$    &$75.900\pm3.824$    &$76.277\pm2.410$                               \cr
        GatedGCN\cite{2017arXiv171107553B}          & $97.34\pm0.14$                          & $67.31\pm0.31$    & $65.667\pm4.899$   &$72.918\pm2.090$    &$76.363\pm2.904$                               \cr
        PNA\cite{corso2020principal}                & $97.94\pm0.12$                          & $70.86\pm0.27$    &-                   &-                   & -                                             \cr
        DGN\cite{beaini2021directional}             & -                                       & $72.84\pm0.42$    &-                   &-                   & -                                             \cr
        EGT\cite{hussain2022global}                 & $98.17\pm0.09$                          & $68.70\pm0.41$    &-                   &-                   & -                                             \cr
        ARGNP\cite{cai2022automatic}                & -                                       & $73.90\pm0.15$    &-                   &-                   & -                                             \cr
        EXPHORMER\cite{shirzad2023exphormer}        & $98.55\pm0.04$                          & $74.69\pm0.13$    &-                   &-                   & -                                             \cr
        \hline
        GNAS-MP\cite{cai2021rethinking}             & $98.01\pm0.10$                          & $70.10\pm0.44$    &-                   &-                   & -                                                                                \cr        
        
    \hline    
        GIG-GatedGCN& $\mathbf{98.80\pm0.03}$          & $\mathbf{77.11\pm0.05}$      &$\mathbf{80.25\pm2.770}$   &$\mathbf{89.09\pm3.151}$      &$\mathbf{85.12\pm1.982}$                          \\
\bottomrule
\end{tabular}
\vspace{10pt}
\begin{tabular}{ccc|cc|c}
    \toprule
    \multicolumn{3}{c}{Graph regression} &\multicolumn{2}{c}{Node classification} & \multicolumn{1}{c}{Edge classification}\cr
    \hline
    ZINC(10K)                   & ZINC-full                                   &AQSOL                   &PATTERN                                 & CLUSTER                                 & TSP\cr
    \multicolumn{3}{c}{MAE $\downarrow$} &\multicolumn{2}{c}{Acc(\%) $\uparrow$} & \multicolumn{1}{c}{F1$\uparrow$}\cr
    \hline
    $0.113\pm0.002$             & $0.278\pm0.003$                             &$1.333\pm0.013$         &$85.61\pm0.03$                          & $69.03\pm1.37$                          & $0.643\pm0.001$\cr
    $0.088\pm0.002$             & $0.387\pm0.015$                             &$1.894\pm0.024$         &$85.59\pm0.01$                          & $64.72\pm1.55$                          & $0.656\pm0.003$\cr
    -                           & $0.042\pm0.003$                             &-                       &-                                       &-                                        &-                \cr
    -                           &$0.045\pm0.006$                              &-                       &-                                       &-                                        &-                \cr
    \hline
    $0.111\pm0.002$             & $0.384\pm0.007$                             &$1.403\pm0.008$         &$78.27\pm0.19$                          & $70.59\pm0.45$                          & $0.671\pm0.002$\cr
    $0.214\pm0.013$             & -                                           &$0.996\pm0.008$         &$85.57\pm0.09$                          & $73.84\pm0.33$                          & $0.838\pm0.002$\cr
    -                           & -                                           &-                       &$86.57\pm0.08$                          & -                                       & -             \cr
    -                           & -                                           &-                       &$86.68\pm0.03$                          & -                                       & -             \cr
    $\mathbf{0.108\pm0.009}$    &-                                            &-                       &$86.82\pm0.02$                          & $79.23\pm0.35$                          & $0.853\pm0.001$\cr
    $0.136\pm0.002$             &-                                            &-                       &-                                       & $77.35\pm0.05$                          & $0.855\pm0.001$\cr
    -                           & -                                           &-                       &$86.74\pm0.02$                          & $78.07\pm0.04$                          & -  \cr
    \hline
    $0.242$                     &-                                            &-                       &$86.80\pm0.10$                          & $62.21\pm0.20$                          & -             \cr        
        
    \hline    
    $0.125\pm0.001$             & $\mathbf{0.029\pm0.001}$    &$\mathbf{0.927\pm0.018}$    &$\mathbf{86.87\pm0.01}$                 & $\mathbf{79.34\pm0.03}$                 & $\mathbf{0.863\pm0.001}$\\
\bottomrule
\end{tabular}
\end{table*}

\begin{table}[t]
\caption{Results on two human skeleton video-based action recognition datasets. Bolder denotes the best performance.}
\label{tab:Skeleton results}
\centering
\begin{tabular}{l|cl}
\toprule
Dataset              & X-Sub & X-View\\
&\multicolumn{2}{c}{Acc(\%)}\\
\hline
ST-GCN\cite{yan2018spatial}& 81.5& 88.3\\
AS-GCN\cite{li2019actional}& 86.8& 94.2\\
SGN\cite{zhang2020semantics}&89.0& 94.5\\
DGNN\cite{shi2019skeleton} &89.9 &96.1\\
ST-TR-agcn\cite{plizzari2021skeleton}&90.3 & 96.3\\
shift-GCN\cite{cheng2020skeleton} &90.7  & 96.5 \\
DC-GCN+ADG\cite{cheng2020decoupling}& 90.8& 96.6\\
Dynamic GCN\cite{ye2020dynamic}& 91.5& 96.0\\
MS-G3D\cite{liu2020disentangling}&91.5&96.2\\
MST-GCN\cite{chen2021multi}& 91.5&96.6\\
CTR-GCN(4-ensemble)\cite{chen2021channel}& 92.4&96.8\\
InfoGCN(4-ensemble)\cite{chi2022infogcn}& 92.7&96.9\\ 
HD-GCN(4-ensemble)\cite{lee2023hierarchically}& 93.0&97.0\\ 
\hline
GIG-ST-GCN   &83.6& 91.4 \\
GIG-CTR-GCN(4-ensemble)    & $\mathbf{93.1}$ & $\mathbf{97.1}$\\
\bottomrule
\end{tabular}
\end{table}

\subsubsection{Evaluation on NTU RGB+D datasets} 
\begin{figure}[h]
\begin{center}
\includegraphics[width=.483\textwidth]{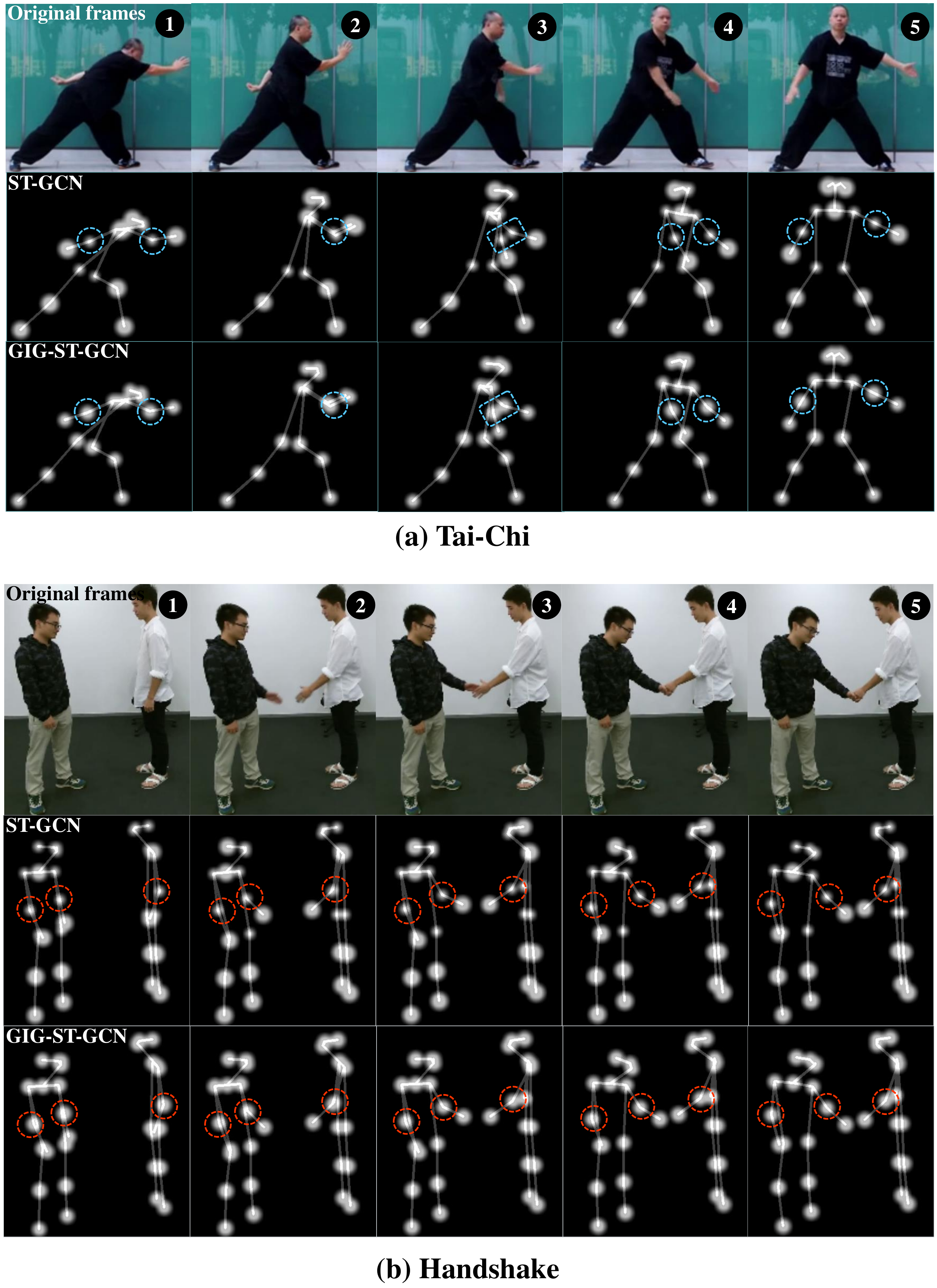}
\end{center}
   \caption{Visualization of features learned by GIG-ST-GCN and ST-GCN, where sizes of white circles represent their importance for action recognition defined by networks. GIG-ST-GCN paid similar attentions to all joints throughout all frames with more attentions to crucial joints than ST-GCN.}
\label{fig:visulization_skeleton}
\end{figure}

Table \ref{tab:Skeleton results} evaluate our GIG on skeleton video (multi-graph data sample)-based action recognition task. Although the advanced CTR-GCN already achieved classification accuracy of $92.4\%$ and $96.8\%$ on X-Sub and X-View datasets, incorporating its vertex and edge updating strategies into our GIG network results in significant (refer to the Supplementary Material for statistical difference analysis) and consistent improvements of $0.8\%$ and $0.3\%$ on X-Sub and X-View datasets, respectively. Specifically, GIG-ST-GCN outperformed the original ST-GCN with more than $2.5\%$ and $3.5\%$ improvements on two datasets, while GIG-CTR-GCN achieved the state-of-the-art performance on both datasets, despite that all competitors were specifically designed for skeleton-based action recognition task while the GIG is a generic GNN that accommodates various tasks. Fig. \ref{fig:visulization_skeleton} highlights the superiority of GIG-ST-GCN over ST-GCN in paying relatively similar attentions to all joints throughout all frames, with greater attention paid to crucial joints (e.g., elbows). In contrast, the original ST-GCN focuses on less important joints (e.g., lower limbs which remained almost still in the action) and ignores crucial joints (e.g., joints moved in the next frame). This indicates our GIG network's ability in exchanging information between GIG vertices (skeleton frames), which considers each human action as a whole-body movements, thereby learned more effective representations.

\begin{table}[ht]
\caption{Experimental results on OGBG-PPA large-scale datasets from \cite{zhao2018work}. Bolder results denote the best performance.}
\label{tab:OGBG-PPA results}
\centering
\begin{tabular}{l|ccc}
    \toprule
    Datasets   & OGBG-PPA    \\
    Models          & Acc(\%)    \\
    \hline
    GPS \cite{rampavsek2022recipe}       &0.8015     \\
    ExpC \cite{yang2020breaking}       &0.7976   \\
    DeeperGCN+FLAG \cite{kong2022robust} &0.7752  \\
    DeeperGCN \cite{li2020deepergcn}  &0.7712  \\
    \hline
    GIG-DeeperGCN  &$\textbf{0.8017}$ \\
    \bottomrule
\end{tabular}
\end{table}

\subsubsection{Evaluation on generic graph analysis datasets}

Table \ref{tab:Benchmark results} and \ref{tab:OGBG-PPA results} compares our GIG with widely-used GNNs on all typical graph analysis tasks (graph/vertex/edge-level analysis), which highlights the following observations: 
\textbf{(i)} GIG achieved state-of-the-art performances on 11/12 datasets, demonstrating its scalability and effectiveness for various graph analysis tasks. While GIG ranked second on the ZINC(10K) dataset, it still largely improves its baseline GatedGCN from 0.214 to 0.125. However, on the ZINC-full dataset that has the same distribution as ZINC(10k) but larger number of samples, our GIG reached out the state-of-the-out results, suggesting the small number of training samples in ZINC(10K) limited the performance of our GIG; \textbf{(ii)} Comparing GIG-GatedGCN with GatedGCN, we found that our GIG-GatedGCN obtained more than $10\%$ average improvement on all datasets. This directly confirms the effectiveness of GIG vertex communication in the proposed GIG, which also maintains the superiority of GatedGCN; and \textbf{(iii)} our GIG outperforms GNAS-MP that employs Neural Architecture Search (NAS) to explore task-specific weights and optimal architecture, showing that GIG can potentially advance the state-of-the-art performance on this benchmark using more advanced vertex/edge updating strategies.

\subsection{Ablation Studies}
\label{subsec:Ablation}
The default vertex/edge updating strategies for generic graph analysis and action recognition ablation experiments are inherited from GatedGCN and ST-GCN, respectively.

\subsubsection{Impacts of individual modules} 
\begin{table}[h]
\caption{Ablation results achieved by \textbf{V1:} treating each GIG vertex of the GIG sample as a vector; \textbf{V2:} combining all GIG vertices as a large graph;  \textbf{V3:} using GVU only;  \textbf{V4:} using GGU only; and  \textbf{V5:} the full GIG network.}
\label{tab:ab_gig}
\centering
\setlength{\tabcolsep}{0.5mm}{
\begin{tabular}{c|ccc|ccc}
\toprule
\multicolumn{3}{c}{Modules} & \multicolumn{3}{c}{Datasets}\cr
\hline
& GSG & GVU & GGU    & MNIST       & TSP   &NTU X-Sub\\
  &   &     &       & Acc(\%)     & $F_1$&Acc(\%)\\
\hline
V1 & \xmark &\xmark    &\xmark   & $26.32\pm0.01$ & $ - $&13.4\\
V2 & \cmark &\xmark    &\xmark   & $94.75\pm0.01$ &$0.853\pm0.001$  &76.3\\
V3 & \cmark &\xmark    &\cmark   & $96.42\pm0.03$ & $0.567\pm0.006$&57.0\\
V4 &\cmark &\cmark    &\xmark   & $97.32\pm0.08$ & $0.835\pm0.003$&81.5\\
V5 &\cmark &\cmark   &\cmark    & $\mathbf{98.80\pm0.03}$ & $\mathbf{0.863\pm0.001}$ &$\mathbf{83.6}$\\    
\bottomrule
\end{tabular}}
\end{table}
Table \ref{tab:ab_gig} indicates that the variant V5, which integrated both GVU and GGU on V2, achieved the highest performance on all tasks, especially better than V1. Here, V1 refers to the system that treats each GIG vertex in a GIG sample as a vector, which is obtained by averaging all graph vertex features in the corresponding GIG vertex, (i.e., a skeleton frame or a graph sample). This finding suggests that \textit{graphs are better vertex representations than vectors for various types of data.} Furthermore, the GVU contributes significantly to the final performance, as evidenced by the improved performance of V5 compared to V3. The results that V5 outperformed V4 also indicate that the GIG network can better model temporal correlations among GIG vertices derived from human action clips, as well as underlying relationship cues among multiple graph samples, thus providing more discriminative information for both action recognition and generic graph analysis tasks. Finally, V5 outperformed V2 which treat all local/global proxy vertices and graph vertices as standard graph vertices, and all edges as standard graph edges, i.e., all vertices are simultaneously processed using GatedGCN/ST-GCN, validated the superiority of our local-global updating strategy (GGU and GVU).

\subsubsection{Impacts of proxy edge number} 
\begin{table}[h]
\centering
\caption{Ablation study for different proxy edge numbers.}
\label{tab:number of proxy edges}
\setlength{\tabcolsep}{1.0mm}{
\begin{tabular}{c|cc|c}
\toprule
    Datasets                  & MNIST                        & PATTERN                    & TSP             \cr
    \hline
    \multirow{1}{*}{\textbf{Proportion}}& \multicolumn{2}{c}{Test Acc(\%) $\uparrow$}  & \multicolumn{1}{c}{Test $F_1\uparrow$}      \cr
    \hline
    10\%                     & $\mathbf{98.80\pm0.03}$      & $\mathbf{86.87\pm0.01}$   & $\mathbf{0.863\pm0.001}$   \cr            
    40\%                     & $98.72\pm0.04$               & $86.83\pm0.02$            & $0.859\pm0.001$   \cr              
    70\%                     & $98.61\pm0.07$               & $86.80\pm0.05$            & $0.860\pm0.003$   \cr            
    100\%                    & $98.71\pm0.05$               & $86.72\pm0.02$            & $0.857\pm0.006$ \cr     
\bottomrule
\end{tabular}}
\end{table}
Table \ref{tab:number of proxy edges} presents the results achieved by connecting each local/global proxy vertex with different proportions of graph vertices, showing that the performance of our GIG is robust to this variable, where the best performance is achieved when each local/global proxy vertex connects 10\% of its graph vertices.

\begin{table}[ht]
\caption{Results achieved by different settings.}
\label{tab:synthesized settings}
\footnotesize
\centering
\setlength{\tabcolsep}{0.5mm}{
    \begin{tabular}{l|cc|c}
    \toprule
        Dataset                  & MNIST                        & PATTERN                    & TSP\cr
        \hline \hline
        \multirow{1}{*}{\textbf{\# GIG layers}}& \multicolumn{2}{c}{Test Acc(\%) $\uparrow$}  & \multicolumn{1}{c}{Test $F_1\uparrow$}\cr
        \hline
        1                        & $97.82\pm0.02$               & $86.79\pm0.01$                    & $0.842\pm0.001$\cr
        2                        & $98.50\pm0.04$                      & $\mathbf{86.87\pm0.01}$           & $0.858\pm0.001$\cr
        3                        & $\mathbf{98.80\pm0.03}$                      & $86.79\pm0.01$                    & $\mathbf{0.863\pm0.001}$\cr
    \hline \hline
    \multirow{1}{*}{\textbf{Proxy edge def.}}& \multicolumn{2}{c}{Test Acc(\%) $\uparrow$}  & \multicolumn{1}{c}{Test $F_1\uparrow$}\cr
    \hline
    Proxy edge def. (i)                     & $\mathbf{98.80\pm0.03}$      & $\mathbf{86.87\pm0.01}$   & $\mathbf{0.863\pm0.001}$   \cr            
    Proxy edge def. (ii)                    & $98.72\pm0.04$               & $86.83\pm0.02$            & $0.859\pm0.001$   \cr              
    Proxy edge def. (iii)                     & $98.61\pm0.07$             & $86.80\pm0.05$            & $0.860\pm0.003$   \cr            
    Proxy edge def. (iv)                    & $98.71\pm0.05$               & $86.72\pm0.02$            & $0.857\pm0.006$ \cr     
    \hline \hline
    \multirow{1}{*}{\textbf{GIG edge def.}}& \multicolumn{2}{c}{Test Acc(\%) $\uparrow$}  & \multicolumn{1}{c}{Test $F_1\uparrow$}\cr
    \hline
    GIG edge def. (i)                     & $98.76\pm0.05$               & $86.82\pm0.03$            & $0.861\pm0.005$   \cr            
    GIG edge def. (ii)                    & $98.78\pm0.04$               & $86.85\pm0.02$            & $0.859\pm0.001$   \cr              
    GIG edge def. (iii)                   & $\mathbf{98.80\pm0.03}$      & $\mathbf{86.87\pm0.01}$   & $\mathbf{0.863\pm0.001}$   \cr   
    \hline \hline
        \multirow{1}{*}{\textbf{Proxy vertex init.}}& \multicolumn{2}{c}{Test Acc(\%) $\uparrow$}  & \multicolumn{1}{c}{Test $F_1\uparrow$}\cr
        \hline
        Random graph vertex     & $98.72\pm0.02$           & $86.84\pm0.01$                          & $0.857\pm0.001$\cr 
        Largest L2 norm   & $98.60\pm0.03$           & $86.49\pm0.02$                          & $0.781\pm0.001$\cr
        Smallest L2 norm  & $98.59\pm0.03$           & $86.48\pm0.01$                          & $0.830\pm0.001$\cr
        Avg. graph vertices          & $\mathbf{98.80\pm0.03}$  & $\mathbf{86.87\pm0.01}$                 & $\mathbf{0.863\pm0.001}$\cr
    \bottomrule
    \end{tabular}}

\end{table}

\subsubsection{Impacts of different proxy edge connection strategies}

Table \ref{tab:synthesized settings} presents the comparison among different proxy edge connection strategies: (i) connecting directed proxy edges from graph vertices that are \textbf{least similar} to each local proxy vertex to it, and each global proxy vertex to graph vertices that are \textbf{most similar} to the corresponding local proxy vertex; (ii) connecting directed proxy edges from graph vertices that are \textbf{least similar} to each local proxy vertex to it, and each global proxy vertex to graph vertices that are \textbf{least similar} to the corresponding local proxy vertex; (iii) connecting directed proxy edges from graph vertices that are \textbf{most similar} to each local proxy vertex to it, and each global proxy vertex to graph vertices that are \textbf{most similar} to the corresponding local proxy vertex; and (iv) connecting directed proxy edges from graph vertices that are \textbf{most similar} to each local proxy vertex to it, and each global proxy vertex to graph vertices that are \textbf{least similar} to the corresponding local proxy vertex. Although there are very small performance variations, the strategy (i) achieved the best performance, which has been utilized in this paper. Further visualization results and analysis can be found in the Supplementary Material. \textit{Note that the generation of proxy edge are only achieved in the GSG module, where every global proxy vertex will be set as zero vector. Then, the similarity calculation mentioned above will actually build on the feature of local proxy vertex.}

\begin{figure}[htbp]
    \centering
    \subfigure[]{ 
        \includegraphics[width=.45\textwidth]{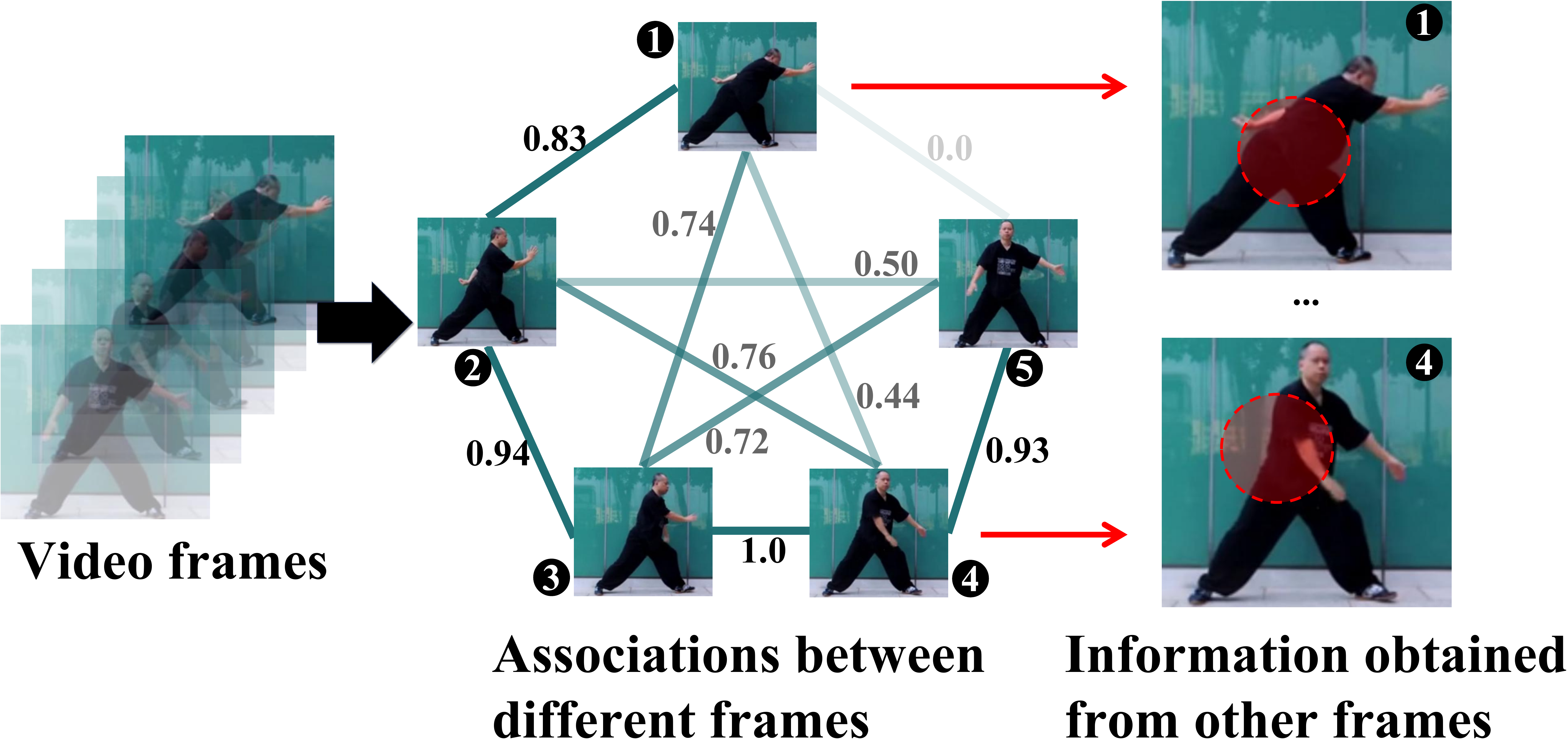}
        \label{fig: edge_feature_vis_hs}
    }
    \hfill 
    \subfigure[]{ 
        \includegraphics[width=.45\textwidth]{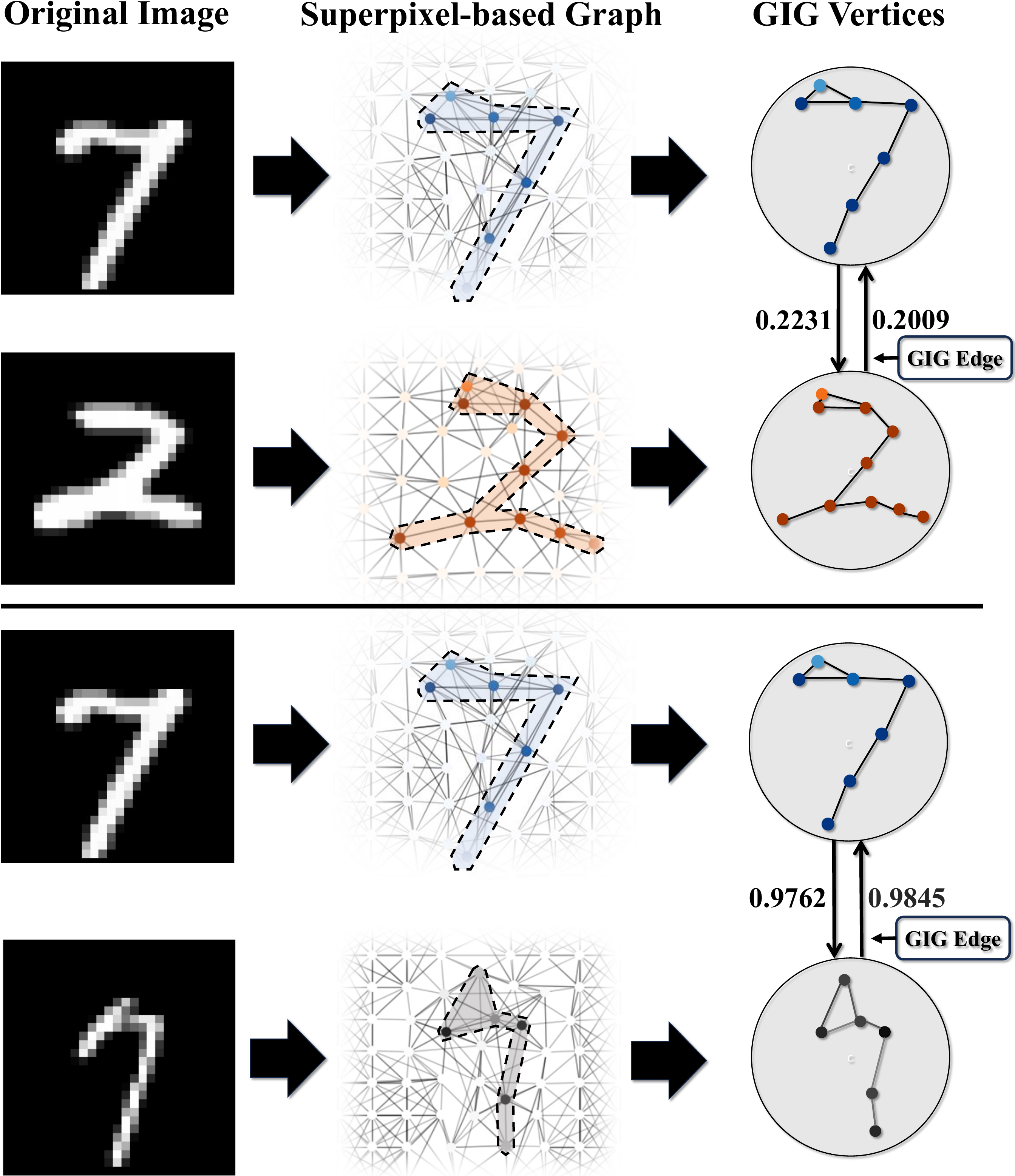}
        \label{fig: edge_feature_vis_mnist}
    }
    \caption{Visualization of: (a) GIG edges constructed for the action recognition tasks; and (b) GIG edges constructed for generic graph analysis tasks.}
\end{figure}
\subsubsection{Impacts of different GIG edge connection strategies} Table \ref{tab:synthesized settings} also compares different GIG edge connection strategies: (i) connecting bidirectional directed edges from each global proxy vertex to local proxy vertices that are \textbf{most similar} to it; (ii) connecting bidirectional directed edges from each global proxy vertex to local proxy vertices that are \textbf{least similar} to it; and (iii) half of bidirectional directed edges are connected from each global proxy vertex to local proxy vertices that are most similar to it, while the rest bidirectional directed edges are connected from each global proxy vertex to local proxy vertices that are \textbf{least similar} to it. Here, we found that the strategy (iii) outperformed other strategies, which is the default setting adopted in the paper. Such results again show that our GIG is robust to this variable. We also illustrate the characteristics of the initial GIG edges learned for skeleton-based action recognition in Figure \ref{fig: edge_feature_vis_hs}, where each connects a pair of GIG vertices (frames). It can be seen that the GIG edges connecting similar frames (i.e., greater temporal proximity) are typically represented by larger weights. Besides, Figure \ref{fig: edge_feature_vis_mnist} further visualizes that GIG edges learned between GIG vertices containing two hand-written numbers (described by superpixel-based graphs/GIG vertices \cite{dwivedi2020benchmarking}) are associated with their similarity. These suggest that the learned GIG edges can partially represent correlations between GIG vertices. 

\subsubsection{Effects of the number of stacked GIG hidden layers} Table \ref{tab:synthesized settings} demonstrates the robustness of our GIG in terms of the number of GIG layers it contains. Although the optimal number of layers and input samples varied across datasets, the performances remains relatively stable, with less than $0.1\%$ variation on vertex classification task when stacking three different numbers of GIG layers. Although even shallow GIG networks outperformed existing deep GNNs (e.g., 2 or 3 layer-GIG outperformed 16 layer-GatedGCN on PATTERN, CLUSTER and TSP), the complexity of GIG networks are practically not more than existing GNNs with the same edge/vertex updating functions (discussed in \ref{subsec:complexity analysis} and the section named \'Runtime Analysis\' in Supplementary Material). Here, one-layer GIG-ST-GCN and GIG-CTR-GCN are used for human skeleton-based action recognition, following their original proposals.

\subsubsection{Effects of different local proxy vertex initialization strategies} Table \ref{tab:synthesized settings} indicates that while the initializing each local proxy vertex by averaging all of its graph vertices yielded highest performances across all three tasks, the differences for four initialization methods were negligible. These suggest that the GSG layer can maintain robust performances across different local proxy vertex initialization methods.

\begin{figure}[htbp]
\begin{center}
\includegraphics[width=.45\textwidth]{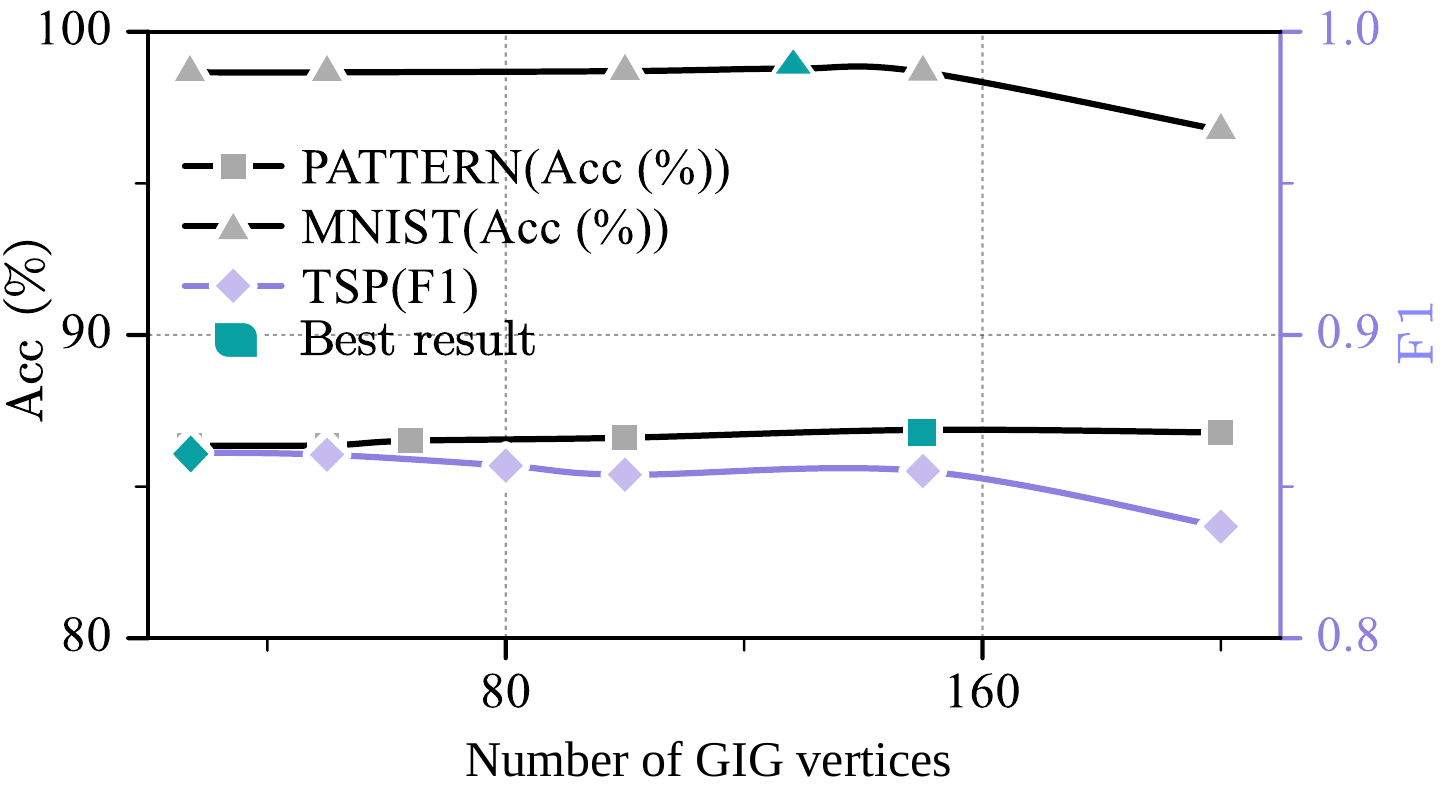}
\end{center}
\caption{Results achieved for GIG samples consisting of different numbers of GIG vertices.}
\label{fig:number-of-graphs}
\end{figure}

\subsubsection{Effects of the GIG vertex number in each GIG sample.} Figure \ref{fig:number-of-graphs} illustrates the  capability of our GIG network in simultaneously handling different numbers of input graph samples for generic graph analysis, which shows that its performances remain stable or slightly varied when the number of input graph samples varies from 27 to 200 on all tasks. This highlights that the potential of the proposed GIG network for real-world applications that require jointly analysing different numbers of graphs.

\begin{table}[ht]
\caption{Results achieved for GIG layers with varying arrangements of stacking GGU and GVU modules.}
\label{tab:ab_order}
\centering
\setlength{\tabcolsep}{1.0mm}{
\begin{tabular}{c|ccc}
\toprule
 Datasets   & MNIST       & TSP   &NTU X-Sub\\
Variants            & Acc(\%)     & $F_1$&Acc(\%)\\
\hline
GGU $\rightarrow$ GVU &$98.59\pm0.02$&$0.855\pm0.001$&57.1\\
GVU $\rightarrow$ GGU &$\mathbf{98.80\pm0.03}$&$\mathbf{0.863\pm0.001}$&$\mathbf{83.6}$\\
\bottomrule
\end{tabular}}
\end{table}

\subsubsection{Effects of GVU and GGU's order in GIG layers.} Table \ref{tab:ab_order} presents a comparison between two variants of the GIG layer, namely locally updating first (i.e., GVU module is placed at the first) and globally updating first (i.e., GGU module is placed at the first). The results demonstrate that the order of updating local and global information has little impact on the performances for the generic graph analysis tasks where no temporal correlations among GIG vertices.

\section{Conclusion and discussion}

This paper proposes a novel GIG network which can process graph-style data (i.e., GIG sample) whose vertices are represented by graphs rather than vectors or single values, allowing to effectively analyze real-world multi-graph data or a batch of graph samples. It also provides flexibility in vertex / edge updating functions by allowing customization from existing GNNs. Experimental results show that: (i) our GIG achieved the state-of-the-art performances on 13 out of 14 benchmark datasets corresponding to different tasks; (ii) our approach is robust to all of its main variables (e.g., proxy edge connection strategies, GIG edge connection strategies, proxy vertex initialization strategies, the number of GIG hidden layers, etc.), validating and emphasizing its generalization capability, i.e., our GIG is easy to be deployed; (iii) the proposed GVU and GGU modules can effectively learn task-related and complementary cues from each graph as well as the relationships among graphs; (iv) Connecting more graph vertices for each local proxy vertex, including more GIG vertices for each global proxy vertex, or adding more GIG layers did not guarantee an absolute improvement in performance, suggesting that the performance of GIG is largely independent to its complexity; and (v) the performance of GIG is basically robust when the order of GVU and GGU is changed.

The main limitation of our approach is that its GSG layer requires an additional task/data-dependent graph representation transformation step to convert multiple non-graph frames/objects contained in real-world data samples into a set of GIG vertices. Besides, the superior performance of our model is at the cost of its relatively high computational complexity despite that practically its runtime is not clearly higher than standard GNNs. As a result, our future work will focus on not only developing an effective and generic graph representation learning module to replace the additional task/data-dependent graph representation transformation step/reducing the model complexity, but also evaluating our GIG on more related tasks/data.

\backmatter

\bmhead{Data Availablity Statement}
The authors affirm that the data supporting the findings of this study are openly accessible at GitHub project \url{https://github.com/graphdeeplearning/benchmarking-gnns} for MNIST, CIFAR10, DD, PROTEINS, ENZYMES, ZINC-10K, ZINC-full, AQSOL, PATTERN, CLUSTER, and TSP, at \url{https://ogb.stanford.edu/docs/graphprop/} for OGBG-PPA, and at \url{https://rose1.ntu.edu.sg/dataset/actionRecognition} for NTU RGB+D. Our code can be accessed at \url{https://github.com/wangjs96/Graph-in-Graph-Neural-Network}.


\bibliography{ref}

\end{document}


\title[Article Title]{Supplementary Material}


\author[1]{\fnm{Jiongshu} \sur{Wang}}\email{jiongshuwang@gmail.com}
\author[3]{\fnm{Jing} \sur{Yang}}\email{y.jing2016@gmail.com}
\author[2]{\fnm{Jiankang} \sur{Deng}}\email{jiankangdeng@gmail.com}
\author[3]{\fnm{Hatice} \sur{Gunes}}\email{Hatice.Gunes@cl.cam.ac.uk}
\author*[1,3]{\fnm{Siyang} \sur{Song}}\email{ss2796@cam.ac.uk}

\affil*[1]{\orgdiv{School of Computing and Mathematical Sciences}, \orgname{University of Leicester}, \orgaddress{\city{Leicester}, \postcode{LE1 7RH}, \state{East Midlands}, \country{United Kingdom}}}

\affil[2]{\orgdiv{Department of Computing}, \orgname{Imperial College London}, \orgaddress{\city{London}, \postcode{SW7 2AZ}, \state{London}, \country{United Kingdom}}}

\affil*[3]{\orgdiv{Department of Computer Science and Technology}, \orgname{University of Cambridge}, \orgaddress{\city{Cambridge}, \postcode{CB3 0FD}, \state{Cambridgeshire}, \country{United Kingdom}}}

\maketitle

\section{Differentiation Analysis of the GIG Layer}
\label{sec:backpropagation}

In this section, we provide the back-propagation analysis of the proposed GIG layer. Supposing that the loss value $\mathcal{L}$ is computed by the objective function $\mathcal{l}$ and $\overline{\mathcal{G}}(\overline{S}_G(\overline{V}_S, \overline{E}_S), \overline{P}_g, \overline{E}_P, \overline{E}_G)$ is the output of the GIG layer, the gradient of the proposed GIG layer w.r.t. the input GIG sample $\mathcal{G}(S_G(V_S, E_S), P_l, P_g, E_P, E_G)$ can be defined as:
\begin{align}
    &\frac{\partial \mathcal{L}}{\partial \mathcal{G}(S_G(V_S, E_S), P_l, P_g, E_P, E_G)} =  \frac{\partial \mathcal{L}}{\partial V_S} + \frac{\partial \mathcal{L}}{\partial E_S} \notag \\
                                                                               &+ \frac{\partial \mathcal{L}}{\partial P_l} + \frac{\partial \mathcal{L}}{\partial P_g} + \frac{\partial \mathcal{L}}{\partial E_P} + \frac{\partial \mathcal{L}}{\partial E_G} 
\end{align}
\noindent There are two types of directed proxy edges, either starting from graph vertices to local proxy vertex or starting from global proxy vertex to graph vertices. Here, we use $E^L_P$ and $E^G_P$ to denote these two kinds of proxy edges, respectively. Besides, $P_l$ represents local proxy vertex and $P_g$ represents global proxy vertex, respectively. Consequently, the $\frac{\partial \mathcal{L}}{\partial E_S}$, $\frac{\partial \mathcal{L}}{\partial V_S}$, $\frac{\partial \mathcal{L}}{\partial P_l}$, $\frac{\partial \mathcal{L}}{\partial P_g}$, $\frac{\partial \mathcal{L}}{\partial E^L_P}$, $\frac{\partial \mathcal{L}}{\partial E^G_P}$ and $\frac{\partial \mathcal{L}}{\partial E_G}$ can be computed as:
\begin{align}
    \frac{\partial \mathcal{L}}{\partial E_S} &= \frac{\partial \mathcal{L}}{\partial \overline{V}}_S\frac{\partial \overline{V}_S}{\partial \hat{V}_S}\frac{\partial \hat{V}_S}{\partial \hat{E}_S}\frac{\partial \hat{E}_S}{\partial E_S} \notag \\
                                              &+\frac{\partial \mathcal{L}}{\partial \overline{V}_S}\frac{\partial \overline{V}_S}{\partial \overline{E}^G_P}\frac{\partial \overline{E}^G_P}{\partial \hat{V}_S}\frac{\partial \hat{V}_S}{\partial \hat{E}_S}\frac{\partial \hat{E}_S}{\partial E_S} \notag \\
                                              &+ \frac{\partial \mathcal{L}}{\partial \overline{V}_S}\frac{\partial \overline{V}_S}{\partial \overline{P}_g}\frac{\partial \overline{P}_g}{\partial \overline{E}_G}\frac{\partial \overline{E}_G}{\partial \hat{P}_l}\frac{\partial \hat{P}_l}{\partial \hat{E}^S_P}\frac{\partial \hat{E}^S_P}{\partial \hat{V}_S}\frac{\partial \hat{V}_S}{\partial \hat{E}_S}\frac{\partial \hat{E}_S}{\partial E_S} \notag \\
                                              &+ \frac{\partial \mathcal{L}}{\partial \overline{V}_S}\frac{\partial \overline{V}_S}{\partial \overline{P}_g}\frac{\partial \overline{P}_g}{\partial \overline{E}_G}\frac{\partial \overline{E}_G}{\partial \hat{P}_l}\frac{\partial \hat{P}_l}{\partial \hat{V}_S}\frac{\partial \hat{V}_S}{\partial \hat{E}_S}\frac{\partial \hat{E}_S}{\partial E_S} \notag \\
                                              &+ \frac{\partial \mathcal{L}}{\partial \overline{V}_S}\frac{\partial \overline{V}_S}{\partial \overline{P}_g}\frac{\partial \overline{P}_g}{\partial \hat{P}_l}\frac{\partial \hat{P}_l}{\partial \hat{E}^S_P}\frac{\partial \hat{E}^S_P}{\partial \hat{V}_S}\frac{\partial \hat{V}_S}{\partial \hat{E}_S}\frac{\partial \hat{E}_S}{\partial E_S} \notag \\
                                              &+ \frac{\partial \mathcal{L}}{\partial \overline{V}_S}\frac{\partial \overline{V}_S}{\partial \overline{P}_g}\frac{\partial \overline{P}_g}{\partial \hat{P}_l}\frac{\partial \hat{P}_l}{\partial \hat{V}_S}\frac{\partial \hat{V}_S}{\partial \hat{E}_S}\frac{\partial \hat{E}_S}{\partial E_S} \notag \\
                                              &+ \frac{\partial \mathcal{L}}{\partial \overline{V}_S}\frac{\partial \overline{V}_S}{\partial  \overline{E}^G_P}\frac{\partial  \overline{E}^G_P}{\partial \overline{P}_g}\frac{\partial \overline{P}_g}{\partial \overline{E}_G}\frac{\partial \overline{E}_G}{\partial \hat{P}_l}\frac{\partial \hat{P}_l}{\partial \hat{E}^S_P}\frac{\partial \hat{E}^S_P}{\partial \hat{V}_S}\frac{\partial \hat{V}_S}{\partial \hat{E}_S}\frac{\partial \hat{E}_S}{\partial E_S} \notag \\
                                              &+ \frac{\partial \mathcal{L}}{\partial \overline{V}_S}\frac{\partial \overline{V}_S}{\partial \overline{E}^G_P}\frac{\partial \overline{E}^G_P}{\partial \overline{P}_g}\frac{\partial \overline{P}_g}{\partial \overline{E}_G}\frac{\partial \overline{E}_G}{\partial \hat{P}_l}\frac{\partial \hat{P}_l}{\partial \hat{V}_S}\frac{\partial \hat{V}_S}{\partial \hat{E}_S}\frac{\partial \hat{E}_S}{\partial E_S} \notag \\
                                              &+ \frac{\partial \mathcal{L}}{\partial \overline{V}_S}\frac{\partial \overline{V}_S}{\partial \overline{E}^G_P}\frac{\partial \overline{E}^G_P}{\partial \overline{P}_g}\frac{\partial \overline{P}_g}{\partial \hat{P}_l}\frac{\partial \hat{P}_l}{\partial \hat{E}^S_P}\frac{\partial \hat{E}^S_P}{\partial \hat{V}_S}\frac{\partial \hat{V}_S}{\partial \hat{E}_S}\frac{\partial \hat{E}_S}{\partial E_S} \notag \\
                                              &+ \frac{\partial \mathcal{L}}{\partial \overline{V}_S}\frac{\partial \overline{V}_S}{\partial \overline{E}^G_P}\frac{\partial \overline{E}^G_P}{\partial \overline{P}_g}\frac{\partial \overline{P}_g}{\partial \hat{P}_l}\frac{\partial \hat{P}_l}{\partial \hat{V}_S}\frac{\partial \hat{V}_S}{\partial \hat{E}_S}\frac{\partial \hat{E}_S}{\partial E_S} 
\end{align}

\begin{table*}[ht]
  \centering
  \caption{Overview of the graph learning datasets involved in this work with their respective prediction levels, prediction tasks, and metrics.}
  \label{tab:dataset_statistics:GA}
  \resizebox{1.\textwidth}{!}{
  \begin{tabular}{l|cccccc}
    \toprule
    Dataset & { \# Graphs } & { Avg. \# nodes } & { Avg. \# edges }  & { Prediction level } & { Prediction task } & { Metric } \\
    \midrule
    ZINC-12K & 12000 & 23.2 & 49.8  & graph & regression & {Mean Abs. Error} \\
    ZINC-full & 250000 & 23.2 & 49.8  & graph & regression & {Mean Abs. Error} \\
    AQSOL & 10000 & 17.6 & 35.8  & graph & regression & {Mean Abs. Error} \\
    \hline
    MNIST & 70000 & 70.6 & 564.5  & graph & {10-class classif.} & Accuracy \\
    CIFAR10 & 60000 & 117.6 & 941.1  & graph & {10-class classif.} & Accuracy \\
    ENZYMES & 600 & 32.63 & 62.14  & graph & {6-class classif.} & Accuracy \\
    PROTEINS &1113 &39.06 &3.70  & graph & {2-class classif.} & Accuracy \\
    D\&D &1,178  &284.32 &5.0  & graph & {2-class classif.} & Accuracy \\
    OGBG-PPA &158,100 &243.4 &2,266.1  & graph & {37-class classif.} & Accuracy \\
    \hline
    PATTERN & 14000 & 117.5 & {4749.2}  & inductive node & {binary classif.} & {Accuracy} \\
    CLUSTER & 12000 & 117.2 & {4301.7}  & graph & {6-class classif.} & Accuracy \\
    \hline
    TSP & 12000 & 275.8 & {6,894.0}  & edge & {6-class classif.} & F1 Score \\
    \bottomrule
  \end{tabular}}
\end{table*}

\begin{table*}[ht]
\caption{Statistics of the two employed human skeleton video-based action recognition datasets.\label{tab:dataset_statistics:SK}}
\centering
    \setlength{\tabcolsep}{1.0mm}{
    \begin{tabular}{l|cccccc}
        \toprule
        Dataset                                      & \# clips                                    & Avg. \# Frames                       & Avg. \# Joints (Nodes)  & { Prediction level } & { Prediction task } & { Metric } \\
        \hline
        NTU X-Sub              &40000                                       &300                    &37.5   & graph & {10-class classif.} & Accuracy \\
        NTU X-View             &38000                                       &300                    &37.5  & graph & {10-class classif.} & Accuracy \\
        \bottomrule
    \end{tabular}}
\end{table*}

\begin{align}
    \frac{\partial \mathcal{L}}{\partial E^L_P} &= \frac{\partial \mathcal{L}}{\partial \overline{V}_S}\frac{\partial \overline{V}_S}{\partial \overline{P}_g}\frac{\partial \overline{P}_g}{\partial \overline{E}_G}\frac{\partial \overline{E}_G}{\partial \hat{P}_l}\frac{\partial \hat{P}_l}{\partial \hat{E}^S_P}\frac{\partial \hat{E}^S_P}{\partial E^L_P}\notag \\
                                              &+ \frac{\partial \mathcal{L}}{\partial \overline{V}_S}\frac{\partial \overline{V}_S}{\partial \overline{P}_g}\frac{\partial \overline{P}_g}{\partial \hat{P}_l}\frac{\partial \hat{P}_l}{\partial \hat{E}^S_P}\frac{\partial \hat{E}^S_P}{\partial E^L_P}\notag \\
                                              &+ \frac{\partial \mathcal{L}}{\partial \overline{V}_S}\frac{\partial \overline{V}_S}{\partial \overline{E}^G_P}\frac{\partial \overline{E}^G_P}{\partial \overline{P}_g}\frac{\partial \overline{P}_g}{\partial \overline{E}_G}\frac{\partial \overline{E}_G}{\partial \hat{P}_l}\frac{\partial \hat{P}_l}{\partial \hat{E}^S_P}\frac{\partial \hat{E}^S_P}{\partial E^L_P}\notag \\
                                              &+ \frac{\partial \mathcal{L}}{\partial \overline{V}_S}\frac{\partial \overline{V}_S}{\partial \overline{E}^G_P}\frac{\partial \overline{E}^G_P}{\partial \overline{P}_g}\frac{\partial \overline{P}_g}{\partial \hat{P}_l}\frac{\partial \hat{P}_l}{\partial \hat{E}^S_P}\frac{\partial \hat{E}^S_P}{\partial E^L_P}\\
    \frac{\partial \mathcal{L}}{\partial E^G_P} &=  \frac{\partial \mathcal{L}}{\partial \overline{V}_S}\frac{\partial \overline{V}_S}{\partial \overline{E}^G_P}\frac{\partial \overline{E}^G_P}{\partial E^G_P} \\
    \frac{\partial \mathcal{L}}{\partial P_l} &= \frac{\partial \mathcal{L}}{\partial \overline{V}_S}\frac{\partial \overline{V}_S}{\partial \overline{P}_g}\frac{\partial \overline{P}_g}{\partial \overline{E}_G}\frac{\partial \overline{E}_G}{\partial \hat{P}_l}\frac{\partial \hat{P}_l}{\partial P_l} \notag \\
                                              &+ \frac{\partial \mathcal{L}}{\partial \overline{V}_S}\frac{\partial \overline{V}_S}{\partial \overline{P}_g}\frac{\partial \overline{P}_g}{\partial \overline{E}_G}\frac{\partial \overline{E}_G}{\partial \hat{P}_l}\frac{\partial \hat{P}_l}{\partial \hat{E}^S_P}\frac{\partial \hat{E}^S_P}{\partial P_l}\notag \\
                                              &+ \frac{\partial \mathcal{L}}{\partial \overline{V}_S}\frac{\partial \overline{V}_S}{\partial \overline{P}_g}\frac{\partial \overline{P}_g}{\partial \hat{P}_l}\frac{\partial \hat{P}_l}{\partial \hat{E}^S_P}\frac{\partial \hat{E}^S_P}{\partial P_l}\notag \\
                                              &+ \frac{\partial \mathcal{L}}{\partial \overline{V}_S}\frac{\partial \overline{V}_S}{\partial \overline{P}_g}\frac{\partial \overline{P}_g}{\partial \hat{P}_l}\frac{\partial \hat{P}_l}{\partial P_l}\notag \\
                                              &+ \frac{\partial \mathcal{L}}{\partial \overline{V}_S}\frac{\partial \overline{V}_S}{\partial \overline{E}^G_P}\frac{\partial \overline{E}^G_P}{\partial \overline{P}_g}\frac{\partial \overline{P}_g}{\partial \overline{E}_G}\frac{\partial \overline{E}_G}{\partial \hat{P}_l}\frac{\partial \hat{P}_l}{\partial P_l}\notag \\
                                              &+ \frac{\partial \mathcal{L}}{\partial \overline{V}_S}\frac{\partial \overline{V}_S}{\partial \overline{E}^G_P}\frac{\partial \overline{E}^G_P}{\partial \overline{P}_g}\frac{\partial \overline{P}_g}{\partial \overline{E}_G}\frac{\partial \overline{E}_G}{\partial \hat{P}_l}\frac{\partial \hat{P}_l}{\partial \hat{E}^S_P}\frac{\partial \hat{E}^S_P}{\partial P_l}\notag \\
                                              &+ \frac{\partial \mathcal{L}}{\partial \overline{V}_S}\frac{\partial \overline{V}_S}{\partial \overline{E}^G_P}\frac{\partial \overline{E}^G_P}{\partial \overline{P}_g}\frac{\partial \overline{P}_g}{\partial \hat{P}_l}\frac{\partial \hat{P}_l}{\partial \hat{E}^S_P}\frac{\partial \hat{E}^S_P}{\partial P_l}\notag \\
                                              &+ \frac{\partial \mathcal{L}}{\partial \overline{V}_S}\frac{\partial \overline{V}_S}{\partial \overline{E}^G_P}\frac{\partial \overline{E}^G_P}{\partial\overline{P}_g}\frac{\partial \overline{P}_g}{\partial \hat{P}_l}\frac{\partial \hat{P}_l}{\partial P_l}\\
    \frac{\partial \mathcal{L}}{\partial P_g} &= \frac{\partial \mathcal{L}}{\partial \overline{V}_S}\frac{\partial \overline{V}_S}{\partial \overline{P}_g}\frac{\partial \overline{P}_g}{\partial P_g} \notag \\
                                              &+ \frac{\partial \mathcal{L}}{\partial \overline{V}_S}\frac{\partial \overline{V}_S}{\partial \overline{E}^G_P}\frac{\partial \overline{E}^G_P}{\partial \overline{P}_g}\frac{\partial \overline{P}_g}{\partial P_g} \\
    \frac{\partial \mathcal{L}}{\partial E_G} &= \frac{\partial \mathcal{L}}{\partial \overline{V}_S}\frac{\partial \overline{V}_S}{\partial \overline{P}_g}\frac{\partial \overline{P}_g}{\partial \overline{E}_G}\frac{\partial \overline{E}_G}{\partial E_G}\notag \\
                                              &+ \frac{\partial \mathcal{L}}{\partial \overline{V}_S}\frac{\partial \overline{V}_S}{\partial \overline{E}^G_P}\frac{\partial \overline{E}^G_P}{\partial \overline{P}_g}\frac{\partial \overline{P}_g}{\partial \overline{E}_G}\frac{\partial \overline{E}_G}{\partial E_G} \\
    \frac{\partial \mathcal{L}}{\partial V_S} &= \frac{\partial \mathcal{L}}{\partial \overline{V}_S}\frac{\partial \overline{V}_S}{\partial \hat{V}_S}\frac{\partial \hat{V}_S}{\partial V_S} + \frac{\partial \mathcal{L}}{\partial \overline{V}_S}\frac{\partial \overline{V}_S}{\partial \hat{V}_S}\frac{\partial \hat{V}_S}{\partial \hat{E}_S}\frac{\partial \hat{E}_S}{\partial V_S}\notag \\
                                              &+ \frac{\partial \mathcal{L}}{\partial \overline{V}_S}\frac{{\partial \overline{V}_S}}{\partial \overline{P}_g}\frac{\partial \overline{P}_g}{\partial \overline{E}_G}\frac{\partial \overline{E}_G}{\partial \hat{P}_l}\frac{\partial \hat{P}_l}{\partial \hat{E}^S_P}\frac{\partial \hat{E}^S_P}{\partial \hat{V}_S}\frac{\partial \hat{V}_S}{\partial V_S}\notag \\
                                              &+ \frac{\partial \mathcal{L}}{\partial \overline{V}_S}\frac{{\partial \overline{V}_S}}{\partial \overline{P}_g}\frac{\partial \overline{P}_g}{\partial \overline{E}_G}\frac{\partial \overline{E}_G}{\partial \hat{P}_l}\frac{\partial \hat{P}_l}{\partial \hat{V}_S}\frac{\partial \hat{V}_S}{\partial V_S}\notag \\
                                              &+ \frac{\partial \mathcal{L}}{\partial \overline{V}_S}\frac{{\partial \overline{V}_S}}{\partial \overline{P}_g}\frac{\partial \overline{P}_g}{\partial \hat{P}_l}\frac{\partial \hat{P}_l}{\partial \hat{E}^S_P}\frac{\partial \hat{E}^S_P}{\partial \hat{V}_S}\frac{\partial \hat{V}_S}{\partial V_S}\notag \\
                                              &+ \frac{\partial \mathcal{L}}{\partial \overline{V}_S}\frac{{\partial \overline{V}_S}}{\partial \overline{P}_g}\frac{\partial \overline{P}_g}{\partial \hat{P}_l}\frac{\partial \hat{P}_l}{\partial \hat{V}_S}\frac{\partial \hat{V}_S}{\partial V_S}\notag \\                 
                                              &+  \frac{\partial \mathcal{L}}{\partial \overline{V}_S}\frac{{\partial \overline{V}_S}}{\partial \overline{E}^G_P}\frac{\partial \overline{E}^G_P}{\partial \hat{V}_S}\frac{\partial \hat{V}_S}{\partial V_S}\notag \\
                                              &+  \frac{\partial \mathcal{L}}{\partial \overline{V}_S}\frac{{\partial \overline{V}_S}}{\partial \overline{E}^G_P}\frac{\partial \overline{E}^G_P}{\partial \hat{V}_S}\frac{\partial \hat{V}_S}{\partial \hat{E}_S}\frac{\partial \hat{E}_S}{\partial V_S}\notag \\ 
                                              &+ \frac{\partial \mathcal{L}}{\partial \overline{V}_S}\frac{{\partial \overline{V}_S}}{\partial \overline{E}^G_P}\frac{\partial \overline{E}^G_P}{\partial \overline{P}_g}\frac{\partial \overline{P}_g}{\partial \overline{E}_G}\frac{\partial \overline{E}_G}{\partial \hat{P}_l}\frac{\partial \hat{P}_l}{\partial \hat{E}^S_P}\frac{\partial \hat{E}^S_P}{\partial \hat{V}_S}\frac{\partial \hat{V}_S}{\partial V_S}\notag \\
                                              &+ \frac{\partial \mathcal{L}}{\partial \overline{V}_S}\frac{{\partial \overline{V}_S}}{\partial \overline{E}^G_P}\frac{\partial \overline{E}^G_P}{\partial \overline{P}_g}\frac{\partial \overline{P}_g}{\partial \overline{E}_G}\frac{\partial \overline{E}_G}{\partial \hat{P}_l}\frac{\partial \hat{P}_l}{\partial \hat{V}_S}\frac{\partial \hat{V}_S}{\partial V_S}\notag \\
                                              &+ \frac{\partial \mathcal{L}}{\partial \overline{V}_S}\frac{{\partial \overline{V}_S}}{\partial \overline{E}^G_P}\frac{\partial \overline{E}^G_P}{\partial \overline{P}_g}\frac{\partial \overline{P}_g}{\partial \hat{P}_l}\frac{\partial \hat{P}_l}{\partial \hat{E}^S_P}\frac{\partial \hat{E}^S_P}{\partial \hat{V}_S}\frac{\partial \hat{V}_S}{\partial V_S}\notag \\
                                              &+ \frac{\partial \mathcal{L}}{\partial \overline{V}_S}\frac{{\partial \overline{V}_S}}{\partial \overline{E}^G_P}\frac{\partial \overline{E}^G_P}{\partial \overline{P}_g}\frac{\partial \overline{P}_g}{\partial \hat{P}_l}\frac{\partial \hat{P}_l}{\partial \hat{V}_S}\frac{\partial \hat{V}_S}{\partial V_S}
\end{align}

\newcommand{\tabincell}[2]{\begin{tabular}{@{}#1@{}}#2\end{tabular}}
\begin{table*}[htbp]
\caption{Runtime analysis of our best GIG-GatedGCN model on seven graph datasets.}
\label{tab:Run-time analysis results:GA}
\centering
    \setlength{\tabcolsep}{1.0mm}{
    \begin{tabular}{l|c|cc|cc|cc|c}
        \toprule
        Model           &Type                                                   & MNIST                                 & CIFAR10                             & ZINC                               & AQSOL                             & PATTERN                           & CLUSTER                           & TSP  \cr
        \hline
        GIG-GatedGCN    & \tabincell{c}{Each iteration\\Each epoch\\Convergence\\Inference}   & \tabincell{c}{0.35s\\2.50m\\3.58h\\0.0026s}    & \tabincell{c}{0.34s\\1.97m\\3.78h\\0.0025s}  &\tabincell{c}{0.21s\\0.28m\\6.69h\\0.0016s}  &\tabincell{c}{0.26s\\0.26m\\0.20h\\0.0025s} &\tabincell{c}{0.60s\\0.67m\\0.48h\\0.0025s} &\tabincell{c}{0.47s\\1.23m\\1.47h\\0.0042s} &\tabincell{c}{0.31s\\1.91m\\1.08h\\0.0090s}\cr
        \hline
        GatedGCN        & \tabincell{c}{Each iteration\\Each epoch\\Convergence\\Inference}   & \tabincell{c}{0.23s\\1.66m\\3.55h\\0.0028s}    & \tabincell{c}{0.20s\\1.19m\\4.15h\\0.0026s}  &\tabincell{c}{0.14s\\0.18m\\0.62h\\0.0017s}  &\tabincell{c}{0.11s\\0.11m\\0.41h\\0.0032s} &\tabincell{c}{0.11s\\0.56m\\3.96h\\0.0024s} &\tabincell{c}{0.23s\\1.12m\\6.81h\\0.0045s} &\tabincell{c}{0.55s\\1.44m\\2.85h\\0.0090s}\cr
        \bottomrule
    \end{tabular}}
\end{table*}

\begin{table*}[htbp]
\caption{Training settings of our best models for 12 generic graph analysis datasets and 2 human-skeleton based action recognition datasets.}
\label{tab:model settings}
\small
\centering
    \setlength{\tabcolsep}{0.5mm}{
    \begin{tabular}{r|c|c|c|c|c}
        \hline
        Dataset      &\#Layer                &Learning rate          & Loss                       & Decay strategy                &Optimizer  \\
        \hline
        MNIST        &3                      &0.0018                 &Cross Entropy               &CosineAnnealing                &AdamW    \\ 
        CIFAR10      &3                      &0.0018                 &Cross Entropy               &CosineAnnealing                &AdamW    \\
        PROTEINS     &3                      &0.0012                 &Cross Entropy               &CosineAnnealing                &AdamW    \\
        D\&D         &3                      &0.0012                 &Cross Entropy               &CosineAnnealing                &AdamW   \\
        ENZYMES      &3                      &0.0012                 &Cross Entropy               &CosineAnnealing                &AdamW    \\
        OGBG-PPA     &1                      &0.01                   &Cross Entropy               &-                              &Adam     \\
        \hline
        ZINC         &3                      &0.0025                 &L1                          &CosineAnnealing                &AdamW    \\
        ZINC-full    &3                      &0.0025                 &L1                          &CosineAnnealing                &AdamW    \\
        AQSOL        &1                      &0.0025                 &L1                          &CosineAnnealing                &AdamW    \\
        \hline
        CLUSTER      &1                      &0.0027                 &Cross Entropy               &ReduceLROnPlateau              &AdamW    \\
        PATTERN      &2                      &0.0030                 &Cross Entropy               &ReduceLROnPlateau              &AdamW    \\
        \hline
        TSP          &3                      &0.002                  &Cross Entropy               &ReduceLROnPlateau              &AdamW    \\
        \hline
        NTU X-Sub    &1                      &0.1                    &Cross Entropy               &MultiStep                      &AdamW    \\
        NTU X-View   &1                      &0.1                    &Cross Entropy               &MultiStep                      &AdamW    \\
        \hline
    \end{tabular}}
\end{table*}

\begin{table*}[htbp]
\caption{Statistical significant difference analysis (T-Test) results between our best GIG systems and the corresponding second best systems for both tasks as well as several ablation studies, where the confidence interval is set to 0.05 (+ denotes there is significant difference while - denotes there is no significant difference).\label{tab:T-test}}
\centering
\setlength{\tabcolsep}{0.5mm}{
    \begin{tabular}{c|c}
    \toprule
    Type                 &Significant difference (P-value)            \\
    \hline
    GIG vs EGT           & + (0.0540)                        \\
    GIG vs InfoGCN           & - (0.1034)                        \\
    GIG vs GatedGCN      & + (0.3238)                         \\
    GIG vs CTR-GCN       & + (0.1682)                         \\
    \hline
    (i) GSG+GVU+GGU vs GSG+GVU & + (0.0280)                      \\
    (ii) Default number of proxy edge vs Connect all graph vertices & + (0.0018) \\
    (iii) Proxy edge default setting vs Most similar connections & + (0.0020)  \\
    (iv) GIG edge default setting vs Most similar connections   & + (0.0003)                \\
    (v) Averaging graph vertices vs Random graph vertex   & + (0.0047)                \\
    (vi) GVU $\rightarrow$ GGU vs GGU $\rightarrow$ GVU     & + (0.2142)                \\
    \bottomrule
    \end{tabular}}
\end{table*}

\section{Model complexity analysis for example GIG-GatedGCN and GIG-GAT networks and runtime Analysis}

A GIG network consists of a GSG layer, a set of GIG hidden layers and a GIG output layer. As discussed in the main document, the overall time complexity of the \textbf{GSG layer} is $O(|\mathcal{G}| \times max(V_S) \times n+|\mathcal{G}|^2 \times n)$, where $|\mathcal{G}|$ represents the number of GIG vertices, $max(V_S)$ indicates the maximal number of graph vertices in all GIG vertices, and $n$ denotes the constant time operation. After that, each \textbf{GIG hidden layer} is made up of a GVU and a GGU module, whose specific time complexity is determined by the employed edge/vertex updating algorithms. 
\begin{itemize}
    \item \textbf{For GIG-GatedGCN}, the total time complexity of the GVU and GGU are both $O(E \times D_e + V \times D_v)$, where $E$ indicates the number of edges in the graph, $D_e$ indicates the dimensionality of the features being processed for edges, $V$ indicates the number of vertices in the graph and $D_v$ indicates the dimensionality of the features being processed for vertices. Consequently, the time complexity of each GIG-GatedGCN hidden layer can be defined as $O(2 \times (E \times D_e + V \times D_v))$.

    \item \textbf{For GIG-GAT}, the total time complexity of the GVU and GGU can be defined as $O(H \times (E \times D_e + V \times D_v))$, where $H$ indicates the number of attention heads, $E$ indicates the number of edges in the graph, $D_e$ indicates the dimensionality of the features being processed for edges, $V$ indicates the number of vertices in the graph and $D_v$ indicates the dimensionality of the features being processed for vertices. Then, the time complexity of each GIG-GAT hidden layer can be defined as $O(2 \times H \times (E \times D_e + V \times D_v))$.

\end{itemize}
The \textbf{GIG output layer} additionally updates graph vertices that are not connected with global proxy vertices as well as graph edges compared to GIG hidden layer. As a result, the total time complexity of each GIG output layer can be defined as $O(E \times D_e + V \times D_v)$ for GIG-GatedGCN and $O(H \times (E \times D_e + V \times D_v))$ for GIG-GAT. As a result, the entire model complexity for a GIG-GatedGCN network that contains $k$ hidden layers can be defined as $O(|\mathcal{G}| \times max(V_S) \times n+|\mathcal{G}|^2 \times n + (2 \times k + 1) \times (E \times D_e + V \times D_v))$.  Meanwhile, the model complexity of a GIG-GatedGCN network that contains $k$ hidden layers can be defined as $O(|\mathcal{G}| \times max(V_S) \times n+|\mathcal{G}|^2 \times n + (2 \times k + 1) \times H \times (E \times D_e + V \times D_v))$.

Table \ref{tab:Run-time analysis results:GA} and Table \ref{tab:Run-time analysis results:SK} further provide the runtime analysis for our GIG networks on various datasets in terms of training and inference, respectively. While our GIG-GatedGCN has more time consumption of each training iteration/epoch than original GatedGCN, it has much less convergence time consumption on five out of seven datasets and very similar inference speed as the original GatedGCN. In addition, the superior performance of GIG-CTR-GCN is at the cost of longer training time. However, its inference speed is not slower than the original CTR-GCN.

\textbf{Influences of the numbers of proxy edges and GIG edges on the GVU and GGU's complexity:} In GVU, the number of proxy edge can be defined as $|E_P^i|/2$. We use $O(f_E^\text{GVU})$ to represent the algorithmic complexity of edge processing involved in the specific algorithm used by the GVU module. Then the time complexity related to the number of proxy edges can be defined as $O(|E_P^i|/2 \times f_E^\text{GVU})$. In GGU, the number of proxy edge is same as before, which can be defined as $|E_P^i|/2$. Continuously, the number of GIG edges can be defined as $|\mathcal{N}_{\text{global}}(P)| \times |P|$. Furthermore, we can use $O(f_E^\text{GGU})$ to represent the algorithmic complexity of edge processing utilised by GGU module. Therefore, there will be time complexity in GGU module as $O(f_E^\text{GGU} \times (|E_P^i|/2 + |\mathcal{N}_{\text{global}}(P)| \times |P|))$ in total. Overall, the main time complexity occupation in both GVU and GGU can be calculated as $O(|E_p^i|/2 \times f_E^\text{GVU} + (|E_P^i|/2 + |\mathcal{N}_{\text{global}}(P)| \times |P|) \times f_E^\text{GGU})$. From information extracted from above, the computational complexity of model during training and inference is closely related to the number of proxy edges and GIG edges. From a large scale of experiments, the strategy we choose to construct connections between local/global proxy vertices and graph vertices is to partially connect each local/global proxy vertex with graph vertices (e.g. 10\% graph vertices of each GIG vertex in this paper) in the corresponding GIG vertex and the strategy utilized to build connections between different global proxy vertices with local proxy vertices from other GIG vertices is upon the KNN algorithm, which is generally no more than 9 neighbours in all downstream tasks. Then the actual extra model complexity related to proxy edge and GIG edge during running can be limited in a small range. In particular, the time complexity discussed above is only related to the FLOPs but all parameters are shared in GVU and GGU modules (e.g., proxy edge and graph edge will utilise same parameters) for different types of edges.

\begin{table}[htbp]
\caption{Runtime analysis of our best GIG-CTR-GCN model on two skeleton video-based action recognition datasets.}
\label{tab:Run-time analysis results:SK}
\centering
    \setlength{\tabcolsep}{0.5mm}{
    \begin{tabular}{c|c|cc}
        \hline
        Model           &Type                                                   & NTU X-Sub                             & NTU X-View             \\
        \hline
        GIG-CTR-GCN    & \tabincell{c}{Each iteration\\Each epoch\\Convergence\\Inference}   & \tabincell{c}{1.20s\\11.37m\\8.72h\\0.0131s}    & \tabincell{c}{1.80s\\16.00m\\15.73h\\0.0152s}  \\
        \hline
        CTR-GCN        & \tabincell{c}{Each iteration\\Each epoch\\Convergence\\Inference}   & \tabincell{c}{0.69s\\7.22m\\7.58h\\0.0135s}    & \tabincell{c}{1.17s\\11.50m\\12.46h\\0.0154s}  \\
        \hline
    \end{tabular}}
\end{table}

\section{Details of the two action recognition datasets}
Table \ref{tab:dataset_statistics:SK} further provides the details of the employed human skeleton video-based action recognition datasets.

\section{Training hyper-parameter settings for all experiments}
We provide detailed training settings for generic graph analysis tasks and skeleton-based video action recognition tasks in Table \ref{tab:model settings} which represent the settings for our best models. Please check the code for more details. All models are optimised using the AdamW \cite{loshchilovdecoupled} optimizer.

\section{Effects of different proportions of similar and non-similar GIG vertices to be connected with each global proxy vertex}
\begin{table}[htbp]
\centering
\caption{Ablation study for different proportions in construction of GIG edges. The (i) represents the proportion of similar GIG vertices adopted and the (ii) represents the proportion of non-similar GIG vertices adopted.}
\setlength{\tabcolsep}{1.0mm}{
    \begin{tabular}{cc|cc|c}
    \hline
        \multicolumn{2}{c}{Strategy}                 & MNIST                        & PATTERN                    & TSP             \cr
        \hline
        \multirow{1}{*}{(i)} &(ii) & \multicolumn{2}{c}{Test Acc(\%) $\uparrow$}  & \multicolumn{1}{c}{Test $F_1\uparrow$}      \cr
        \hline
        10\%       &90\%              & $98.76\pm0.04$               & $86.80\pm0.05$            & $0.857\pm0.001$   \cr 
        20\%       &80\%              & $98.75\pm0.03$               & $86.81\pm0.03$            & $0.860\pm0.001$   \cr 
        30\%       &70\%              & $98.77\pm0.03$               & $86.83\pm0.02$            & $0.859\pm0.002$   \cr            
        40\%       &60\%                & $98.75\pm0.04$               & $86.85\pm0.01$            & $0.859\pm0.001$   \cr              
        50\%       &50\%               & $\mathbf{98.80\pm0.03}$          & $\mathbf{86.87\pm0.01}$          & $\mathbf{0.863\pm0.001}$   \cr            
        60\%       &40\%               & $98.79\pm0.01$               & $86.83\pm0.03$            & $0.862\pm0.001$ \cr     
        70\%       &30\%              & $98.73\pm0.05$               & $86.82\pm0.01$            & $0.862\pm0.003$   \cr            
        80\%       &20\%                & $98.79\pm0.03$               & $86.85\pm0.02$            & $0.859\pm0.001$   \cr
        90\%       &10\%              & $98.71\pm0.07$               & $86.79\pm0.03$            & $0.857\pm0.006$ \cr 
        \hline
    \end{tabular}
}
\label{tab:different proportions of GIG edge}
\end{table}

Table \ref{tab:different proportions of GIG edge} presents the ablation studies as comparing the different combination of proportions of similar and non-similar GIG edges. The results show that 50\% similar GIG edges and 50 \% non-similar
global-edges possesses the best performance.

\section{Statistical difference analysis}

We provide statistical significant difference analysis results in Table \ref{tab:T-test}, where we first compare our best GIG-based system with the second best systems for seven generic graph analysis tasks and two human skeleton-based video action recognition tasks, respectively. It can be seen that our GIG significantly better than the second best system (i.e., EGT) in generic graph analysis while also achieving promising p-value results compared to the InfoGCN on skeleton-based action recognition. Furthermore, the p-value results for GIG and its baselines, including GatedGCN and CTR-GCN, show the significant difference as well.

Meanwhile, it also compares our best systems with the second best variants for six ablation study experiments, which are: (i) the impacts of individual modules in GIG; (ii) different number of proxy edges; (iii) proxy edge definition strategies; (iv) GIG edge definition strategies; (v) local proxy vertex initialization strategies; and (vi) the impacts of the order of different modules.

\begin{table}[ht]
\centering
\caption{Results achieved for different similarity calculation strategies used for measuring the similarity between: (i) the local proxy vertex and its corresponding graph vertices for defining proxy edges; and (ii) local proxy vertices and global proxy vertices for defining GIG edges.}
\setlength{\tabcolsep}{0.5mm}{
    \begin{tabular}{c|cc|c}
    \hline
        \multicolumn{1}{c}{Strategy}                 & MNIST                        & PATTERN                    & TSP             \cr
        \hline
        -                   & \multicolumn{2}{c}{Test Acc(\%) $\uparrow$}  & \multicolumn{1}{c}{Test $F_1\uparrow$}      \cr
        \hline
        L1 norm                     & $98.64\pm0.03$               & $86.57\pm0.09$            & $0.854\pm0.003$   \cr 
        L2 norm                    & $98.76\pm0.05$               & $86.81\pm0.06$            & $0.861\pm0.002$   \cr 
        Cosine similarity          & $\mathbf{98.80\pm0.03}$               & $\mathbf{86.87\pm0.01}$            & $\mathbf{0.863\pm0.001}$   \cr         
        
        \hline
    \end{tabular}
}
\label{tab:different strategy for construction of proxy edge and GIG edge}
\end{table}

\section{Analysis of different similarity calculation strategies}

\noindent Table \ref{tab:different strategy for construction of proxy edge and GIG edge} compares the results achieved by using three different metrics to measure the similarity between (i) the local proxy vertex and its corresponding graph vertices for defining proxy edges; and (ii) local proxy vertices and global proxy vertices for defining GIG edges. It is clear that the employed cosine similarity achieved marginal but consistent advantages over others, i.e., our approach is relatively robust to the employed similarity measurement.


\bibliography{ref}